\documentclass[lettersize,journal]{IEEEtran}
\usepackage{amsmath,amsfonts}
\usepackage{algorithm}
\usepackage{array}
\usepackage[caption=false,font=normalsize,labelfont=sf,textfont=sf]{subfig}
\usepackage{textcomp}
\usepackage{stfloats}
\usepackage{url}
\usepackage{verbatim}
\usepackage{graphicx}
\usepackage{cite}
\usepackage{diagbox} 
\usepackage{amssymb}
\usepackage{algpseudocode}
\usepackage{graphicx}
\usepackage{xcolor}
\usepackage{cases}
\usepackage{multirow}
\usepackage{multicol}
\usepackage{adjustbox}
\usepackage{makecell}
\usepackage[perpage]{footmisc}
\usepackage{hyperref}
\begin{document}

\title{CrossFi: A Cross Domain Wi-Fi Sensing Framework Based on Siamese Network}

\author{Zijian Zhao,\IEEEmembership{} Tingwei Chen,\IEEEmembership{} Zhijie Cai,\IEEEmembership{} Xiaoyang Li,\IEEEmembership{} Hang Li\IEEEmembership{}, Qimei Chen\IEEEmembership{}, Guangxu Zhu\IEEEmembership{}
\thanks{
This work was supported in part by National Natural Science Foundation of China (Grant No. 62371313), in part by ShenzhenHong Kong-Macau Technology Research Programme (Type C) (Grant No. SGDX20230821091559018), in part by Guangdong Major Project of Basic and Applied Basic Research under Grant 2023B0303000001, in part by Shenzhen Science and Technology Program under Grant JCYJ20220530113017039, in part by the National Key R\&D Program of China under Grant 2024YFB2908001, in part by the Key Research and Development Plan of Hubei Province under Grant 2023BCB041, in part by the Wuhan Science and Technology Achievement Transformation Project under Grant 2024030803010178. (Corresponding Author: Guangxu Zhu)}
\thanks{Zijian Zhao is with Shenzhen Research Institute of Big Data, Shenzhen 518115, China, and also with the School of Computer Science and Engineering, Sun Yat-sen University, Guangzhou 510275, China (e-mail: zhaozj28@mail2.sysu.edu.cn).}
\thanks{Tingwei Chen, Zhijie Cai, Xiaoyang Li, Hang Li, and Guangxu Zhu are with the Shenzhen Research Institute of Big Data, The Chinese University of Hong Kong (Shenzhen), Shenzhen 518115, China (e-mail:  tingweichen@link.cuhk.edu.cn; zhijiecai@link.cuhk.edu.cn; lixiaoyang@sribd.cn; hangdavidli@sribd.cn; gxzhu@sribd.cn).}
\thanks{Qimei Chen is with the School of Electronic Information, Wuhan University, Wuhan, 430072, China (e-mail: chenqimei@whu.edu.cn).}
\thanks{Copyright (c) 20xx IEEE. Personal use of this material is permitted. However, permission to use this material for any other purposes must be obtained from the IEEE by sending a request to pubs-permissions@ieee.org.}
}


\markboth{IEEE INTERNET OF THINGS JOURNAL}%
{Shell \MakeLowercase{\textit{et al.}}: A Sample Article Using IEEEtran.cls for IEEE Journals}


\maketitle

\begin{abstract}
In recent years, Wi-Fi sensing has garnered significant attention due to its numerous benefits, such as privacy protection, low cost, and penetration ability. Extensive research has been conducted in this field, focusing on areas such as gesture recognition, people identification, and fall detection. However, many data-driven methods encounter challenges related to domain shift, where the model fails to perform well in environments different from the training data. One major factor contributing to this issue is the limited availability of Wi-Fi sensing datasets, which makes models learn excessive irrelevant information and over-fit to the training set. Unfortunately, collecting large-scale Wi-Fi sensing datasets across diverse scenarios is a challenging task. To address this problem, we propose CrossFi, a siamese network-based approach that excels in both in-domain scenario and cross-domain scenario, including few-shot, zero-shot scenarios, and even works in few-shot new-class scenario where testing set contains new categories.
The core component of CrossFi is a sample-similarity calculation network called CSi-Net, which improves the structure of the siamese network by using an attention mechanism to capture similarity information, instead of simply calculating the distance or cosine similarity. Based on it, we develop an extra Weight-Net that can generate a template for each class, so that our CrossFi can work in different scenarios.
Experimental results demonstrate that our CrossFi achieves state-of-the-art performance across various scenarios. In gesture recognition task, our CrossFi achieves an accuracy of 98.17\% in in-domain scenario, 91.72\% in one-shot cross-domain scenario, 64.81\% in zero-shot cross-domain scenario, and 84.75\% in one-shot new-class scenario. 
The code for our model is publicly available at \href{https://github.com/RS2002/CrossFi}{https://github.com/RS2002/CrossFi}.
\end{abstract}

\begin{IEEEkeywords}
Siamese Network, Cross-domain Learning, Few-shot Learning, Zero-shot Learning, Wi-Fi Sensing, Channel Statement Information
\end{IEEEkeywords}

\section{Introduction}

Recently, Integrated Sensing and Communications (ISAC) has emerged as a prominent and popular technology direction aimed at enhancing the efficiency and intelligence of communication systems. Wi-Fi, as one of the key technologies in the realm of Internet of Things (IoT) communications, has found widespread application in various settings such as homes, offices, and public spaces \cite{zhu2023pushing}. In addition to its role in facilitating communication, Wi-Fi also holds promise as a sensing tool, owing to its characteristics including privacy protection, affordability, and penetration capability. In passive Wi-Fi sensing \cite{passive-survey,chen2021distributed}, by leveraging the variations in signal strength and multipath propagation caused by different objects and actions, it is possible to extract valuable information like Channel Statement Information (CSI) and Received Signal Strength Indicator (RSSI) to sense the environment and detect specific actions.

Wi-Fi sensing has attracted significant research attention, particularly in areas such as fall detection \cite{falldewideo,fall}, gesture recognition \cite{SN-MMD,CSI-BERT}, and people localization \cite{lofi,yu2024complex}. These advancements have demonstrated the immense potential of Wi-Fi sensing in domains such as elderly care, military applications, and medical fields. However, a major challenge faced by existing Wi-Fi sensing models lies in their limited robustness. Even a slight change in the environment can lead to a significant deterioration in model performance or even complete failure. Addressing this critical issue is crucial for the practical deployment and utilization of Wi-Fi sensing devices.

Current Wi-Fi sensing methods can be categorized into two types: model-based methods \cite{wang2015understanding} and data-driven methods \cite{RatioFi}. Model-based methods require significant expertise and extract different signal features for different tasks. However, these methods are challenging to design and often have low accuracy, particularly in complex Wi-Fi sensing scenarios. Moreover, most of these methods are not easily transferable to other tasks. On the other hand, data-driven methods, with deep learning as a prime example, can address these challenges by directly learning from the data, without any explicit assumption on the underlying model.

However, data-driven Wi-Fi sensing methods face a significant challenge in cross-domain scenarios \cite{chen2023cross}. The Wi-Fi signal is highly influenced by the environment, making models trained in specific environments ineffective when applied to new environments. While collecting large amounts of data in diverse environments might seem like an intuitive solution, acquiring signal data is much more challenging compared to other modalities such as images or text. This is because signal data always requires specialist equipment to collect, and there is a lack of rich resources available on the web. Additionally, the data format of the signal is device-dependent, making it nearly impossible to utilize signal data from different public datasets simultaneously. To address this issue, it is crucial to develop a robust framework that can be applied across different environments with minimal modifications.

\begin{figure}
\centering 
\includegraphics[width=0.5\textwidth]{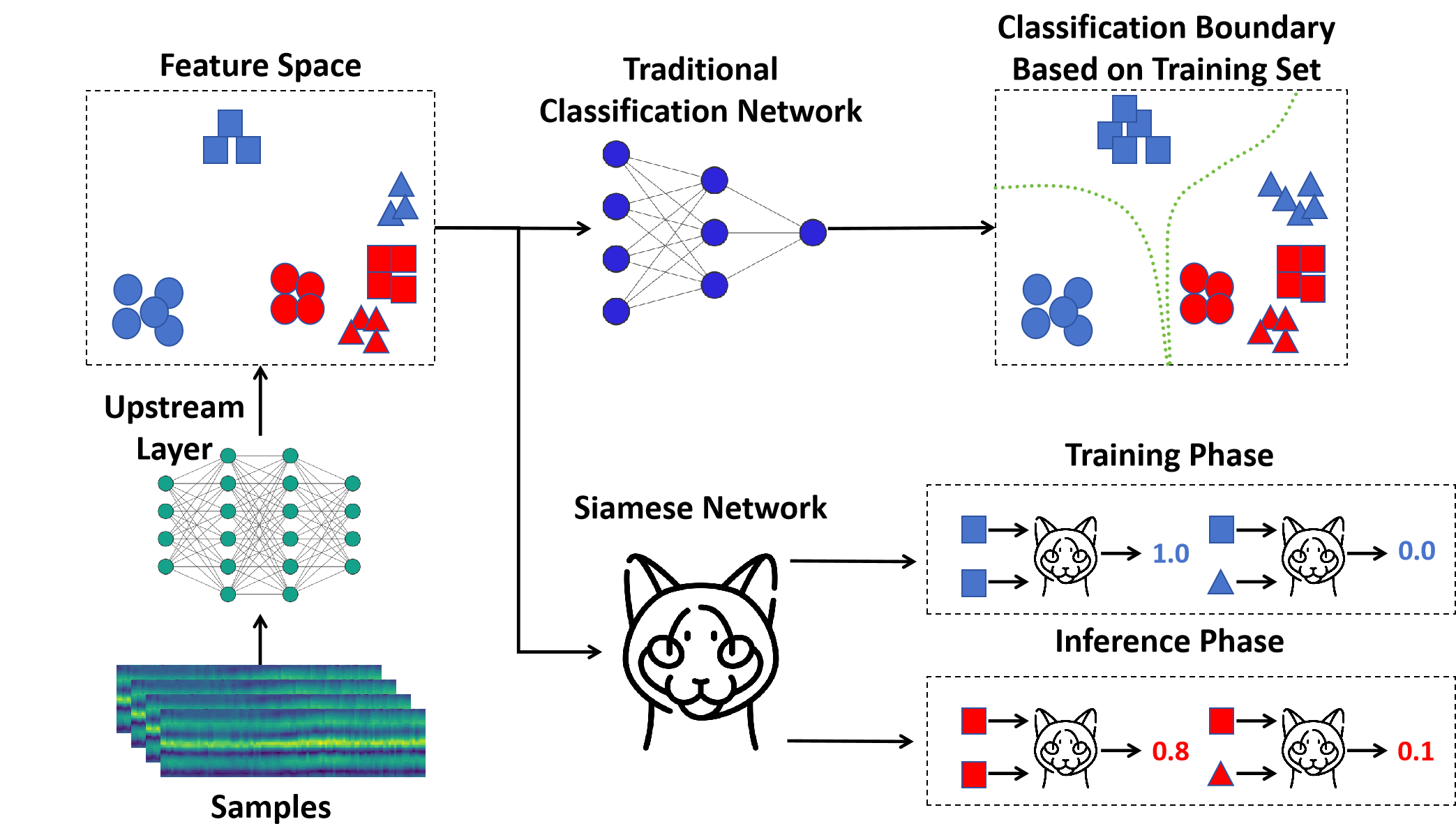}
\caption{Comparison Between Siamese Network and Traditional Classification Network: Different shapes represent different categories. The blue and red items represent samples from the source domain and target domain, respectively. The green line represents the classification boundary. They remain consistent across the following figures.}
\label{siamese}
\end{figure}

Several research studies have focused on the cross-domain topic in Wi-Fi sensing and machine learning \cite{AutoFi,Airfi,Wione}. Among these, the siamese network \cite{siamese} has been proven to be an efficient method. For traditional neural networks, the upstream layers capture the feature embedding of the input, and the downstream layers realize the specific tasks like classification based on it. Shown as Fig. \ref{siamese}, when there is a significant gap between the distribution of training data and testing data, which corresponds to the scenario of cross-domain tasks, the embedding distribution between them also has a huge difference. This can lead to a significant decrease in performance or even failure of the downstream classifier. In contrast, the siamese network calculates the similarity or distance (referred to as ``similarity" for simplicity) of embeddings from the upstream layers between two samples instead of directly outputting the classification result. By this approach, even though the embedding distribution of the target domain may not be similar to the source domain, the model can still capture the similarity relationship between samples from the same domain, which has been proved by many works \cite{koch2015siamese}. By employing the idea of comparative learning, it can capture more information between positive and negative pairs. With its structure, the siamese network has a unique advantage in one-shot scenarios.

\begin{figure*}
\centering 
\includegraphics[width=\textwidth]{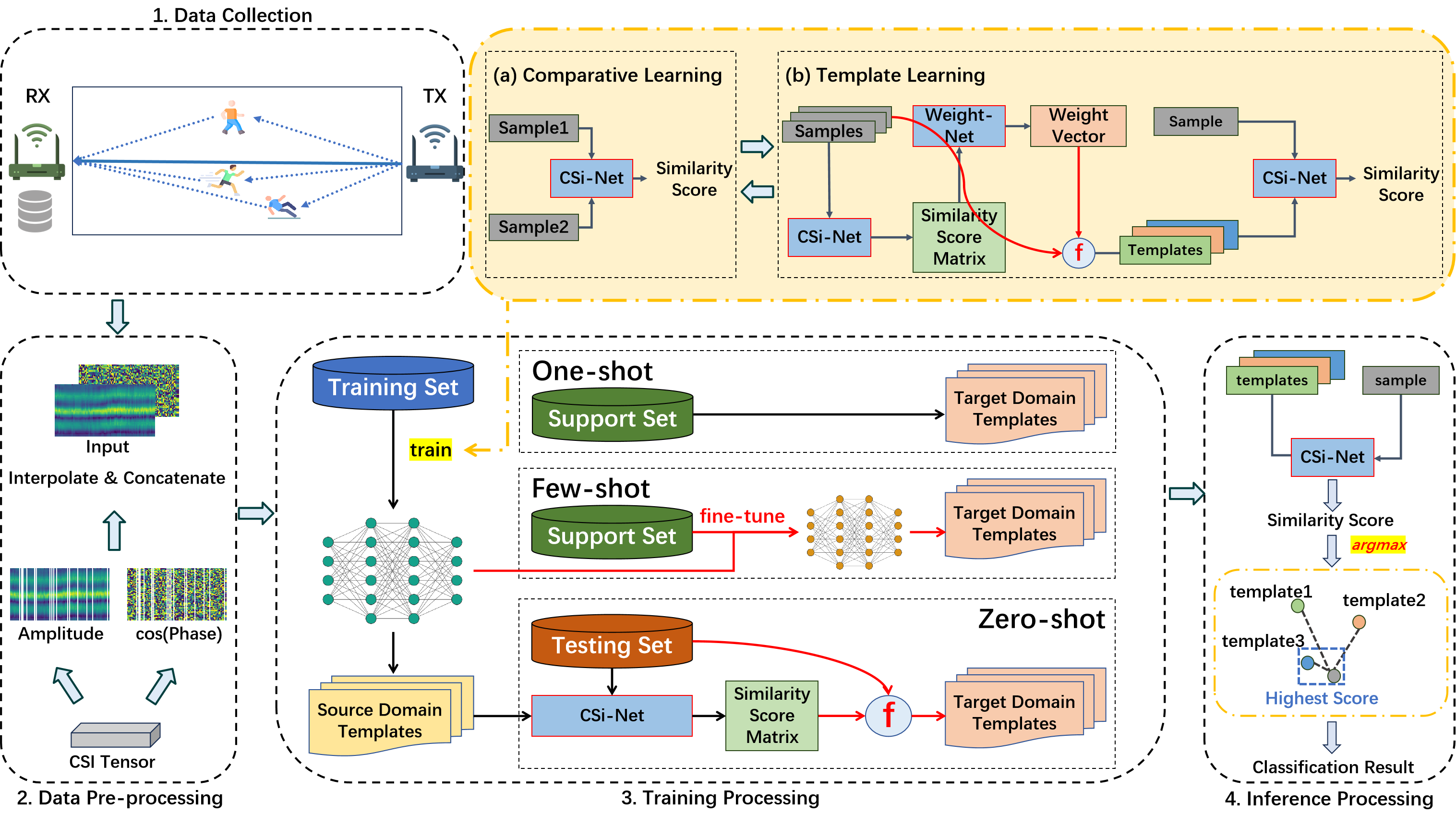}
\caption{Workflow: Our model can be organized into four main phases: data collection, data pre-processing, training, and inference. The training phase encompasses two stages, namely comparative learning and template learning. The red chapter `f' represents a function for template generation, which is weighted average operation in in-domain and few-shot scenarios and argmax operation in zero-shot scenario. It remains consistent in Fig. \ref{Weight-Net}.}
\label{workflow}
\end{figure*}

However, in the traditional siamese network, the similarity between different samples is evaluated by computing the Gaussian distance \cite{siamese} or cosine similarity \cite{siamese-survey} between the embeddings of the two samples, which may not capture enough relationship between samples' feature. Therefore, we propose an attention-based method to calculate the similarity within the network, where we call the improved siamese network as \underline{C}ross Domain \underline{Si}amese \underline{Net}work (CSi-Net). Furthermore, as the siamese network solves the cross-domain task well in one-shot tasks, we hope to extend its success to more scenarios. As a result, in each scenario, we design a corresponding template generation method for each category and, during inference, identify the most similar template for each sample as its category result. Specifically, we propose a Weight-Net to generate templates based on the relationship between different samples adaptively. The whole workflow of our method is shown as Fig. \ref{workflow}, called CrossFi. We evaluate our model on the WiGesture dataset \cite{CSI-BERT} for cross-domain and new-class gesture recognition and people identification tasks. The experimental results demonstrate that our model achieves the most advanced performance in most scenarios.








In summary, the main contributions of this work are:

\textbf{(1) CrossFi -- A Universal Framework for Cross Domain Wi-Fi Sensing:} Aiming at cross-domain Wi-Fi sensing, we propose a universal framework called CrossFi that can work in in-domain, few-shot cross-domain, zero-shot cross-domain scenarios, and even few-shot new-class scenario where testing set contains new class samples not present in training set. 
The framework consists of two components: CSi-Net, which is used to calculate the similarity between samples, and Weight-Net, which is used to generate templates for each class in the source domain and target domain, respectively. During inference, CSi-Net can classify samples by calculating the similarity between them and each template.

\textbf{(2) CSi-Net -- Similarity Calculator:} In view of the low information usage of traditional siamese networks, we propose CSi-Net to improve its structure. To this end, we design an attention mechanism-based method to allow the model extract similarity information through a learning process, rather than directly calculating the distance or cosine similarity.

\textbf{(3) Weight-Net --  Adaptive Template Generator:} To extend siamese network to other scenarios beyond one-shot setting, we design a Weight-Net to generate templates of each class for classification. It uses the similarity matrix output by CSi-Net to identify the quality of samples, which can then be used as the mixing ratio to generate templates by weighted averaging samples. Instead of randomly selecting samples as templates to imitate the one-shot scenario, our Weight-Net provides high-quality templates, which can improve the model's performance.

\textbf{(4) Experiment Evaluation:} We evaluate our model's performance in different in-domain, cross-domain and new-class scenarios on a public dataset, for gesture recognition and people identification tasks. The results confirm the superiority of the proposed method. We also use a series of ablation studies to prove the efficiency of our modifications to the siamese network and our template generation method.

The rest of this paper is structured as follows: Section \ref{Related Work} introduces previous few-shot learning and zero-shot learning methods. Section \ref{Wi-Fi Sensing Basics} provides the basic principles of CSI and Wi-Fi sensing. Section \ref{Methodology} introduces our method structure and system workflow according to different scenarios in detail. Section \ref{Experiment} presents comparative experiments and ablation studies to demonstrate the superiority of our method. Section \ref{IoT Deployment} introduces a real-time system based on our method, highlighting its potential for practical Internet of Things (IoT) applications. Section \ref{Discussion} examines the advantages and limitations of the proposed method. Finally, Section \ref{Conclusion} concludes the paper and points to potential directions for further research.


\section{Related Work} \label{Related Work}
For better understanding, we first provide a description of the scenarios discussed in this paper. According to the domain distribution of training set and testing set, we can divide Wi-Fi sensing task into in-domain Wi-Fi sensing, where the data domains in the training set and testing set are the same, and cross-domain Wi-Fi sensing, which can be further split based on the availability of data in the source and target domains. In the few-shot (also known as k-shot) scenario, the training set consists of a large amount of data from the source domain and only a few data from the target domain. Specifically, in the k-shot scenario, there are only $k$ samples available in the training set for each class in the target domain. To clarify the problem further, we refer to this subset of data as the support set and the remaining data as the training set. In the one-shot scenario, $k$ is equal to $1$. Finally, in the zero-shot scenario, the training set consists entirely of data from the source domain, while the testing set consists entirely of data from the target domain. 
Additionally, in this paper, we also investigate the few-shot scenario in the context of a new category task, where the testing set is from the same domain as the training set, but includes new classes. We provide $k$ samples for each new category in the training set.

Currently, most cross-domain Wi-Fi sensing methods can be divided into three types. \textbf{(1) The domain-invariant feature extraction method} \cite{widar} aims to extract features of CSI independent of the domain. However, this method always requires extensive experimental knowledge and some prerequisites. When the task or prerequisites change, significant effort is needed to redesign the feature extractor. \textbf{(2) The data generation method} \cite{he2023ai} seeks to synthesize samples in the target domain. Unlike image and text data, evaluating the quality of generated samples in this context is challenging. Current methods mostly rely on the accuracy in downstream tasks and the confusion degree of the discriminator for evaluation, but they have limitations in certain contexts. \textbf{(3) The domain adaptation method} \cite{Fewsense} seeks to transfer the knowledge learnt from source domain to the target domain. This approach is seen as the most promising solution in cross-domain Wi-Fi sensing, due to its high performance, versatility, robustness, and low workload compared to the above two methods  \cite{chen2023cross}. Therefore, we focus on domain adaptation method in this paper. Depending on whether labeled data from target domain is available at the training phase, two major scenarios, i.e., few-shot learning and zero-shot learning are considered in the literature as surveyed below.


\subsection{Few-shot Learning}
Most research on cross-domain Wi-Fi sensing focuses on the few-shot scenario, particularly the one-shot scenario, which is a special case within few-shot learning. The siamese network \cite{siamese} has emerged as a powerful framework widely utilized in this area \cite{SN-MMD,CNN-based,Wione}, either directly or indirectly. 
Current few-shot learning-based Wi-Fi sensing methods can be divided into two types: contrastive learning methods like the siamese network and clustering methods like the prototypical network \cite{snell2017prototypical}. Among them, most research focuses on contrastive learning methods, where there are many similar or variant structures to the siamese network, such as the matching network \cite{vinyals2016matching}, deep similarity evaluation network \cite{hu2021wigr}, and triplet network \cite{triplet}. Additionally, several other works, although not directly utilizing the siamese network, exhibit similar overall frameworks. Specifically, Yin et al. \cite{Fewsense} and Shi et al. \cite{siamese-cos} replaced the similarity metric of Gaussian distance with cosine similarity.

Furthermore, some works in Wi-Fi sensing \cite{siamese-full1,siamese-full2} have extended the siamese network to the in-domain scenario and demonstrated superior performance. This paper further expands this method to the zero-shot scenario, presenting a unified framework applicable to all scenarios and demonstrating the best performance.

\subsection{Zero-shot Learning}
Currently, there are few studies on zero-shot scenarios in the field of Wi-Fi sensing. Even Airfi \cite{Airfi} realized it by introducing the domain generalization method, it requires multi-domain information in the training set, which cannot always be satisfied. In the field of machine learning, the most popular methods can be divided into two types. The first type focuses on designing appropriate loss functions to ensure that the conditional probability in the source domain and target domain are the same \cite{MK-MMD,MMD}. The second type is based on network design like Domain Adversarial Neural Networks (DANN) \cite{DANN}, which first employ neural networks to extract domain-independent features and then train classifiers on these domain-invariant features. For example, Shu et al. \cite{curriculum-dann} combined DANN with curriculum learning, while Yu et al. \cite{global-local} proposed a method based on local domain adversarial adaptation and global domain adversarial adaptation. However, both types of methods have their own limitations. In the case of loss function-based methods, achieving an exact match between the conditional probabilities learned in the source domain and the target domain is challenging (as demonstrated in a simple proof in Section \ref{few-shot}). This is because the labels in the target domain are inaccessible during training, leading to an unknown data distribution in each category. As for DANN-based methods, while extracting domain-invariant features, the feature extractor may unintentionally ignore some important features related to the classification task, resulting in low classification accuracy in both the source and target domains. Our experiments in Section \ref{Zero-shot Experiment} demonstrate that traditional zero-shot learning methods do not perform well in Wi-Fi sensing tasks. Furthermore, both methods share the common disadvantage of requiring raw data from testing set during the training phase, which is not always feasible. In contrast, the method proposed in this paper overcomes this limitation.

In addition to these two types of methods, there are also other representative approaches. Pinheiro \cite{center-template} proposed a method similar to the siamese network \cite{siamese}, but trained the feature extractor separately on the source domain and target domain. Additionally, templates based on the centers of samples from the source domain were employed for each category. However, due to the significant domain gap, the templates from the source domain may not perform well in the target domain. Saito et al. \cite{confidience-level} introduced a novel method that generates pseudo-labels for target domain samples based on confidence levels computed by the classifier trained on the source domain. Similarly, due to the domain gap, the pseudo-labels may have low accuracy because the classifier trained on the source domain often fails to perform well on the target domain.

\section{Wi-Fi Sensing Basics} \label{Wi-Fi Sensing Basics}
CSI is utilized to provide feedback on the characteristics of a wireless channel. Considering a scenario where both the transmitter and receiver are situated within the same indoor space, the transmitted signal traverses multiple paths, experiencing reflections, refractions, or scattering, before reaching the receiver. Mathematically, this channel can be represented as:
\begin{equation}
Y=HX+N,
\end{equation}
where $Y$ and $X$ are the matrices of the received and transmitted signals, respectively, $N$ is the vector of noise signals, and $H$ represents the channel matrix.

The channel frequency response can be represented as:
\begin{equation}
    H(f,t)=H_s(f,t)+H_d(f,t),
\end{equation}
where $f$ is the subcarrier frequency, and $t$ is the time-domain sampling point. The equation can be divided into $H_s(f,t)$, the static component, and $H_d(f,t)$, the dynamic component. The CSI tensor includes dimensions for the number of transmit and receive antennas. However, since our setup is that each transmitter and receiver has one antenna, these dimensions can be disregarded.

In tasks such as gesture recognition and identification, different gestures and the extent of individual movements cause variations in the dynamic component. By capturing these variations, we can predict actions and identify people through CSI.

\section{Methodology} \label{Methodology}
\subsection{Overview}



The workflow of our CrossFi is shown in Fig. \ref{workflow}, encompassing four main phases: data collection, data pre-processing, training, and inference. The method is realized by two neural networks: CSi-Net, which is used to calculate the similarity between samples, and Weight-Net, which is used to evaluate the sample quality and further to generate template for each class. For each scenario, we design a proper training method respectively. In the inference phase, users can use the trained CSi-Net and generated template to realize the final classification.

In this section, we first provide a detailed introduction to the networks structure in Section \ref{Model Structure}. Subsequently, we describe the whole workflow in Section \ref{Workflow Description}. Finally, we discuss the different designs of the training process in different scenarios, addressed in Section \ref{Key Scenarios}.

\subsection{Model Structure}\label{Model Structure}
\begin{figure}[!t]
\centering 
\includegraphics[width=0.38\textwidth]{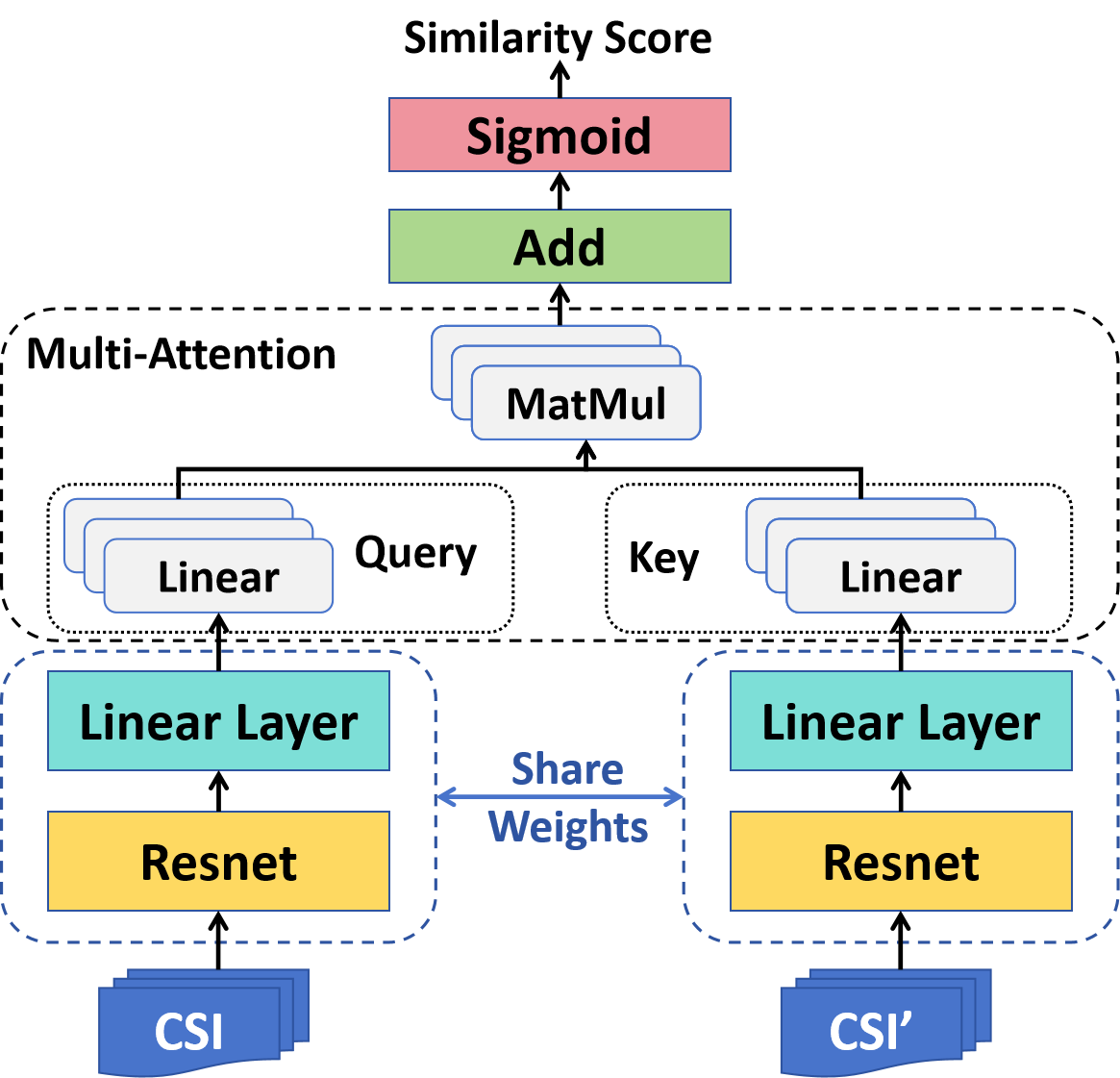}
\caption{Architecture of CSi-Net: CSi-Net utilizes ResNet as a feature extractor and employs a multi-attention mechanism to compute similarity.}
\label{CSi-Net}
\end{figure}

\begin{figure}[!t]
\centering 
\includegraphics[width=0.38\textwidth]{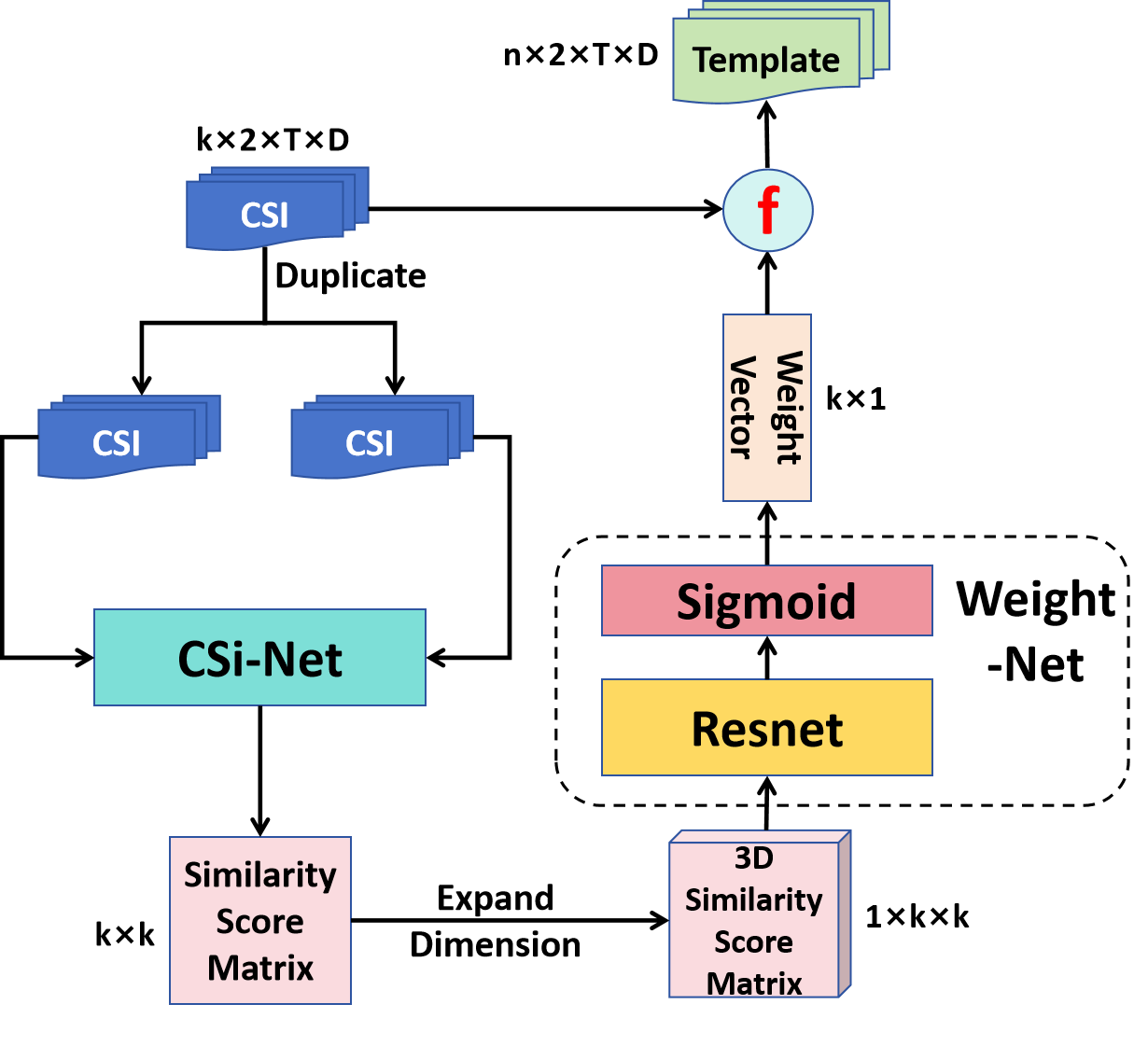}
\caption{Illustration of the template generation method . The proposed Weight-Net is presented within the dashed box. Here, $k, t, D, n$ represent the sample number, packet number, number of sub-carriers across all antennas, and class number, respectively.}
\label{Weight-Net}
\end{figure}

In this section, we describe the structure of CSi-Net and Weight-Net in detail. As depicted in Fig. \ref{CSi-Net}, CSi-Net is based on a traditional Siamese network \cite{siamese}, where two twin networks with tied parameters serve as bottom feature extractors, and the top layer combines the extracted features for similarity calculation. Differently, CSi-Net enhances the original structure by incorporating a multi-attention layer \cite{Transformer} to assess the similarity between two inputs, rather than relying solely on computing the distance between the output of the final model layer. This modification is motivated by the remarkable performance of attention mechanisms in capturing relationships between objects across various domains \cite{attention}. However, unlike the traditional attention mechanism, we only utilize the ``query'' and ``key'' components, omitting the  ``value'' component, which means we only employ the attention matrix as shown in Eq. \ref{attention}:
\begin{equation}
\begin{aligned}
\text{Multi-Attn} (q,k) & = \frac{1}{h} \sum_{i=1}^h (q W_i^Q + b_i^Q) (k W_i^K+ b_i^K)^T \ ,\\
S & = \text{Sigmoid}(\frac{\text{Multi-Attn} (q,k)}{t}) \ ,
\label{attention}
\end{aligned}
\end{equation}
where $h$ represents the number of attention heads, $q \in R^{b_1,d_1}, k \in R^{b_2,d_1}$ represent the outputs from two branches of the bottom embedding layers, which we call the query branch and key branch, $W_i^Q, W_i^K \in R^{d_1,d_2}, b_i^Q, b_i^K \in R^{d_2}$ represent the weights and biases of each linear layer in the query layer and key layer, $t$ represents the temperature which can control the smooth level of sigmoid, $S \in R^{b_1,b_2}$ represents the similarity score matrix, $b_1, b_2$ represent the batch sizes of the query branch and the key branch, $d_1, d_2$ represent the output dimensions of the bottom embedding layers and linear layers in the multi-attention layer. By this mechanism, when there are $b_1$ samples input to the query branch and $b_2$ samples input to the key branch, the model will output a similarity score matrix $S$ of size $b_1 \times b_2$, representing the similarity between each sample pair from the two branches. The attention mechanism, which essentially involves matrix multiplication, offers the same computational complexity as computing Gaussian Distance while yielding more valuable information.

Here, we do not make the parameters of the query layer and weight layer shared as the bottom twin networks. The original idea of parameter sharing is to keep the symmetry of the network, which means the order of the input pair does not affect the output. However, in our method, we introduce extra templates that are generated, not real samples. When computing the similarity between real samples and templates, we input real samples to the query branch and templates to the key branch. By using dependent parameters in the query layer and key layer, we can make the key layer have better capacity to extract features of the template while not influencing the performance of the query layer, which can focus on the feature extraction of real samples.

Furthermore, we still keep the parameters shared in the bottom twin networks, which is viewed as common feature extractors. We choose ResNet \cite{ResNet} as the feature extractor due to its excellent performance in numerous previous wireless sensing works \cite{ResNet-CSI1, ResNet-CSI2}. Inspired by transfer learning \cite{pan2020transfer}, we also use the pre-trained parameters of ResNet from image tasks, which can help increase the model's convergence speed and improve its performance. To align the structure of CSI with ResNet, we represent each CSI sample as $c_i \in \mathbb{R}^{2 \times t \times D}$, where $i$ represents the sample index, $t$ denotes the number of CSI samples in each sample, and $D$ represents the number of sub-carriers across all antennas. Each CSI sample consists of two channels, corresponding to the amplitude and phase, respectively. We also modify the first convolution layer of ResNet to adapt to the 2-channel input.


Then we introduce Weight-Net through template generation process, shown as Fig. \ref{Weight-Net}. The network is a ResNet with an extra sigmoid function in top layer. In template generation process, some samples are first duplicated and input to the two branches of CSi-Net. Then Weight-Net takes the similarity score matrix output by CSi-Net as input and output an weight vector, which represents the quality of each samples. The intuition of using similarity score to evaluate sample quality is that: if a sample has high similarity to some samples and low similarity to others, it has a high quality; if a sample has similar similarity to all samples, it may include too much noise and has a low quality. After getting the weight vector, a function `f' will use the original input samples and weight vector to generate templates. In in-domain and few-shot scenario, `f' represents weighted average operation, which takes the weight vector as the weight for each sample. And in zero-shot scenario, `f' represents argmax operation that we choose the samples with the largest weight within each class as the template. We will introduce the reason of using different `f' in Section \ref{Key Scenarios}.

\subsection{Workflow Description}\label{Workflow Description}
\subsubsection{Data Collection and Data Pre-processing} 
After collecting data from different domains, we split them as training set, support set, and testing set. In training phase, we will use the labeled training set and support set and unlabeled testing set, which is optional. In the data pre-processing phase, we initially compute the amplitude and phase of the CSI since the network cannot directly process complex CSI data. Next, we calculate the cosine value of the phase, as the original phase values may exhibit discontinuities between $-\pi$ and $\pi$, which can impact the network's performance. Finally, we employ an interpolation method to fill in any missing CSI positions, ensuring consistent data dimensions are maintained.


\subsubsection{Training Phase}
As shown in the yellow part of Fig. \ref{workflow}, the training phase includes two alternate steps: comparative learning, where we train the CSi-Net to evaluate the similarity of two samples, and template learning, where we train the Weight-Net for template generation and CSi-Net to evaluate the similarity between samples and templates simultaneously. Since the input of CSi-Net is different in these two steps, we choose to execute them alternately instead of in order, to ensure the model can calculate the similarity both within samples and between samples and templates, and avoid catastrophic forgetting.

In the comparative learning step, we follow the same approach as the traditional siamese network, utilizing the loss function shown in Eq. \ref{comp loss}:
\begin{equation}
\begin{aligned}
L_{com} = & \sum_{i,j} l_{com}^{i,j} \ , \\
l_{com}^{i,j} = & \alpha \mathbf{1}\{\text{label}(c_i)=\text{label}(c_j')\} (1-S_{i,j})^2\\
& +  \mathbf{1}\{\text{label}(c_i) \neq \text{label}(c_j')\} S_{i,j}^2 \ , \\
S_{i,j} = & \text{CSi-Net} (c_i, c_j') \ , 
\label{comp loss}
\end{aligned}
\end{equation}
where $c_i,c_j'$ represent the $i^{th}$ sample in the query branch and the $j^{th}$ sample in the key branch, $S_{i,j}$ represents the similarity between $c_i$ and $c_j'$ calculated by the CSi-Net, $\text{label}(c_i)$ indicates the category to which $c_i$ belongs, $\alpha$ is a weight factor for positive pairs to solve the long-tail problem, as the number of positive pairs is usually significantly fewer than negative pairs, and the function $\mathbf{1}$ is the indicator function that returns 1 when the condition is true and 0 otherwise.


In the template learning phase, the Weight-Net first generates templates for each class, which will be introduced in the next section. Then we still use the comparative learning method to train the Weight-Net and CSi-Net simultaneously, incorporating the loss function shown in Eq. \ref{temp loss}: 
\begin{equation}
\begin{aligned}
L_{tem} = & \sum_{i,j} l_{tem}^{i,j} \ , \\
l_{tem}^{i,j} = & \alpha \mathbb{1}\{\text{label}(c_i)=j\} (1-S_{i,j})^2\\
& + \mathbb{1}\{\text{label}(c_i) \neq j\} S_{i,j}^2 \ , \\
S_{i,j} = & \text{CSi-Net} (c_i, T_j) \ , 
\label{temp loss}
\end{aligned}
\end{equation}
where $T \in R^{n,2,t,D}$ represents the template matrix, $n$ represents class number, $T_j$ represents the template of class $j$, and $S_{i,j}$ represents the similarity between sample $c_i$ and template $T_j$.

We can abstract the processing of the template learning step as follows:
\begin{equation}
\begin{aligned}
s&=\text{f}(x,x;\theta_c) \ , \\
t&=\text{g}(s;\theta_t) \ , \\
\hat{y}&=\text{f}(x',t;\theta_c) \ , \\
L&=\text{l}(\hat{y},y) \ ,
\label{temp process}
\end{aligned}
\end{equation}
where f, g represent CSi-Net and Weight-Net, $\theta_c,\theta_t$ are their parameters, $x,x'$ are CSI samples, $s$ is the similarity score, $t$ is the template, $y,\hat{y}$ are the ground truth and prediction result, and l, $L$ are the loss function and its value. Then the gradient descent can be executed according to the partial derivative of the network parameters:
\begin{equation}
\begin{aligned}
\frac{\partial L}{\partial \theta_t} &= \frac{\partial L}{\partial \hat{y}} \frac{\partial \hat{y}}{\partial t} \frac{\partial t}{\partial \theta_t} \ , \\
\frac{\partial L}{\partial \theta_c} & = \frac{\partial L}{\partial \hat{y}} (\frac{\partial \text{f}(x',t;\theta_c)}{\partial \theta_c} + \frac{\partial \text{f}(x',t;\theta_c)}{\partial t} \frac{\partial \text{g}(s;\theta_t)}{\partial s} \frac{\partial \text{f}(x,x;\theta_c)}{\partial \theta_c})  \ .
\label{gradient}
\end{aligned}
\end{equation}

\subsubsection{Inference Phase}
In the inference phase, we can calculate the similarity score matrix between the testing samples and the template of each class, using the trained CSi-Net and the generated templates in the target domain. The classification result is given by Eq. \ref{classification}:
\begin{equation}
\begin{aligned}
S_{i,j} = & \text{CSi-Net} (c_i, T_j) \ , \\
Y_i = & \underset{j}{\text{argmax}} (S_{i,j}) \ , 
\label{classification}
\end{aligned}
\end{equation}
where $Y_i$ is the classification result of sample $c_i$.

\subsection{Key Scenarios}\label{Key Scenarios}


\subsubsection{In-domain}
Even though the in-domain scenario is not the highlight of this paper, we first introduce how our model works in this scenario as it serves as the base for the few-shot and zero-shot scenarios. Moreover, the experiment results in the next section also show that our model outperforms other in-domain Wi-Fi sensing methods significantly.

The main challenge in the in-domain scenario is to find the optimal template for each category. Previous template generation methods in siamese networks often randomly select samples from the training set or simply take the average of the training set as the template \cite{SiFi}. However, these approaches do not capture the full range of features for each category. To address this limitation, we use Weight-Net to generate templates adaptively.

In detail, we first randomly select $k$ samples from the training set and calculate their weight vector following Fig. \ref{Weight-Net}. By combining the weighted average of the selected samples according to their corresponding weights, we generate the adaptive templates. Weight-Net dynamically determines the weight by analyzing the quality of each sample using the similarity matrix. Since the source domain and target domain are the same in the in-domain scenario, the generated template can be used for both the training and testing sets. The template generation algorithm is outlined in Algorithm \ref{full-shot template}.

\begin{algorithm}
  \caption{Template Generation in In-domain Scenario}
    \textbf{Require}: \\
    \hspace*{1em} $\text{CSi-Net}(\cdot, \cdot)$  \\
    \hspace*{1em} $\text{Weight-Net}(\cdot)$  \\
    \hspace*{1em} Training Set: $C^s$ \\
    \hspace*{1em} Sample Num: $k$ \\
    \textbf{Return}: \\
    \hspace*{1em} Template: $T \in R^{n,2,t,D}$ 
  \begin{algorithmic} [1]
    \State Sample k CSI samples from $C^s$: $C' \sim \text{Sample}(C^s,k)$
    \State Calculate similarity matrix: $S := \text{CSi-Net}(C', C')$
    \State Calculate weights: $W := \text{Weight-Net}(S)$
    \State Initialize $T\in R^{n,2,t,D}$ and $T_w \in R^{n}$ as zero matrices
    \For{$i = 1$ to $k$}
         \State $T[\text{label}(C'[i])] := T[\text{label}(C'[i])] + W[i] C'[i]$
         \State $T_w[\text{label}(C'[i])] := T_w[\text{label}(C'[i])] + W[i]$
    \EndFor
    \State Calculate template $T:=\frac{T}{T_w}$
    \\ \Return  $T$
  \end{algorithmic}
\label{full-shot template}
\end{algorithm}

\subsubsection{Few-shot} \label{few-shot}
In the few-shot learning scenario, we observe that the method used in the in-domain scenario does not perform well due to the presence of a domain gap. The template generated by the source domain cannot be directly applied to the target domain. As a result, we employ the pre-train and fine-tune method, which is very common in cross-domain methods. First, we use the method from the in-domain scenario to pre-train the CSi-Net and Weight-Net. Then, we fine-tune the model using the support set, which consists of limited labeled samples in the target domain. We use the template generated in the support set as the final template for inference. Furthermore, in Section \ref{Effect of Multi-Attention Module}, we have discovered that the traditional similarity measuring method, Gaussian distance, outperforms the multi-attention mechanism in this cross-domain scenario. Therefore, we employ the Gaussian distance in the few-shot scenario of the cross-domain task.

\begin{figure}
\centering 
\includegraphics[width=0.45\textwidth]{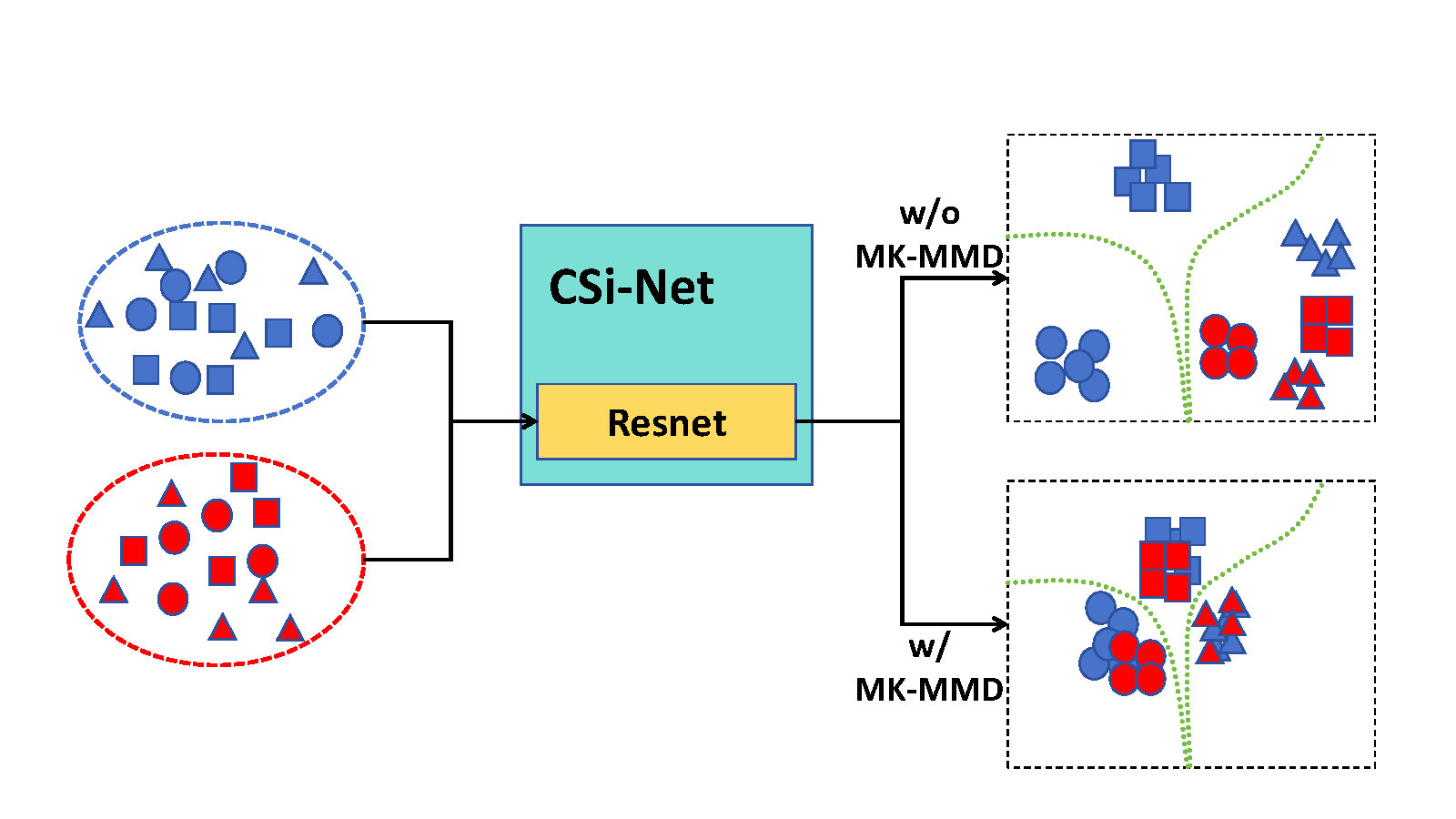}
\caption{Effect of MK-MMD: The blue and red points represent samples from the source domain and target domain, respectively. The green line indicates the classification boundary. MK-MMD helps align the data distributions of the source and target domains in feature space, thereby enabling the classification boundary to function effectively in the target domain.}
\label{MK-MMD effect}
\end{figure}

To ensure better transferability and generalization of the model, it is desirable for the output of the ResNet to exhibit a similar distribution regardless of whether the input comes from the source domain or the target domain. The Multiple Kernel Variant of Maximum Mean Discrepancies (MK-MMD) method can measure the difference between the distributions of two domains \cite{MK-MMD}. Therefore, to minimize the MK-MMD between the source domain and the target domain, we introduce a loss function based on MK-MMD, as shown in Eq. \ref{MK-MMD}, which will be added to both the comparative loss in Eq. \ref{comp loss} and the template-based loss in Eq. \ref{temp loss}.
\begin{equation}
\begin{aligned}
& L_{MK-MMD} = \sum_{j} \beta_j \text{MMD}_j^2(C^s, C^t) \ , \\
& \text{MMD}_j^2(C^s, C^t) \\
& = || \frac{\sum_{c^s_i \in C^s} \phi_j (\text{ResNet} (c^s_i))}{|C^s|} - \frac{\sum_{c^t_i \in C^t} \phi_j (\text{ResNet} (c^t_i))}{|C^t|} ||^2 \ , 
\label{MK-MMD}
\end{aligned}
\end{equation}
where $C^s$ and $C^t$ represent the sample sets from the source domain and target domain respectively, $|\cdot|$ denotes the size of the set, $||\cdot||$ represents the L2 norm, $\phi_j(\cdot): R^D \rightarrow H$ is a function that maps from the real number space to the Reproducing Kernel Hilbert Space (RKHS), $\beta_j$ is a hyper-parameter that satisfies $\sum_j \beta_j=1$. The main purpose of using the MK-MMD loss is to narrow the distance between the source domain and the target domain, which helps improve the efficiency of the feature extractor as shown in Fig. \ref{MK-MMD effect}.

As mentioned before, we found that MK-MMD does not seem to directly help the classifier in the field of Wi-Fi sensing (refer to the experiment result in Section \ref{Zero-shot Experiment}). Here we use a simple proof to illustrate that it cannot be guaranteed that $P_s(y|x;\theta)=P_t(y|x;\theta)$ if we only make $P_s(x)=P_t(x)$. However, this is what MK-MMD does. The proof is shown in Eq. \ref{MKMMD problem}.
\begin{equation}
\begin{aligned}
P(y|x)&=\frac{P(x|y)P(y)}{P(x)} \ , \\
P_s(y|x)&=P_t(y|x)\frac{P_s(x|y)P_s(y)}{P_s(x)}\frac{P_t(x)}{P_t(x|y)P_t(y)}\\
&=P_t(y|x)\frac{P_t(x)}{P_s(x)}\frac{P_s(x|y)}{P_t(x|y)}\frac{P_s(y)}{P_t(y)} \ ,\\
\label{MKMMD problem}
\end{aligned}
\end{equation}
where $P_s$ and $P_t$ represent the probabilities in the source domain and target domain, respectively. $x$, $y$, and $\theta$ denote the input data, label, and network parameters.

\begin{figure}
\centering 
\includegraphics[width=0.45\textwidth]{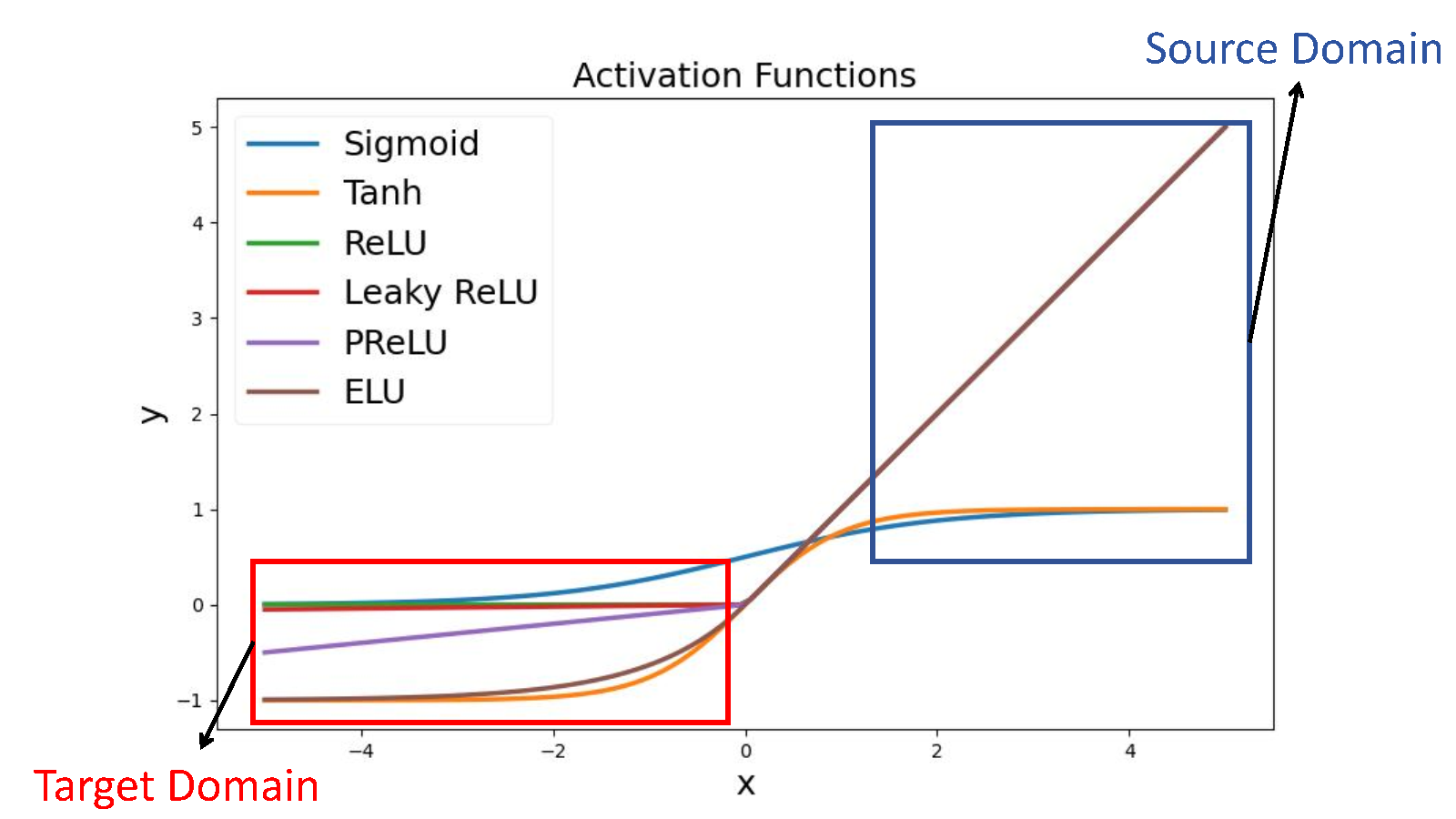}
\caption{Domain Shift Problem: This figure illustrates a potential failure caused by domain shift. If all the data in the target domain falls within the saturated region of the activation function, it renders the subsequent layers of the network ineffective (i.e. regardless of the inputs, the outputs will be similar).}
\label{activation}
\end{figure}

However, despite the lack of direct impact on the classifier, we have observed that both in our work and in a previous siamese network-based method \cite{siamese-cos}, MK-MMD does improve the overall model performance. We believe this improvement is attributed to the structure of the siamese network, which does not directly provide the classification result but computes similarity based on the embedding result. MK-MMD helps the bottom feature extractor obtain better features by minimizing the discrepancy between $P_t(x)$ and $P_s(x)$. This behavior is similar to the effect of batch normalization in networks, which addresses the problem of covariate shift. For instance, in activation functions, if the source domain data predominantly lie in the positive half-shaft while the target domain data predominantly lie in the negative half-shaft, as illustrated in Fig. \ref{activation}, the model struggles to extract meaningful features from the target domain. Particularly for the ReLU function, all data in the negative half-shaft are mapped to 0, resulting in the model losing information about the similarity of the target domain.

What's more, it is important to note that the availability of data in the target domain is not always guaranteed. Therefore, the use of the MK-MMD loss is an optional component in our model. Nonetheless, we can still use the MK-MMD loss between the training set and the support set. In Section \ref{Few-shot Experiment}, we will demonstrate that our model can still function without the MK-MMD loss, albeit with lower performance.

In particular, in the one-shot scenario, there are very few samples in the support set, and regardless of the output of the Weight-Net, the template remains the same throughout because there is only one sample in each category. As a result, the fine-tuning step may not improve the model performance but only increase the training time. Therefore, we follow the same method as in traditional siamese networks, where during inference, the template for each category in the target domain consists of the individual samples belonging to that category.

\subsubsection{Zero-shot}
In the zero-shot scenario, we cannot obtain a template as we do in the few-shot scenario. Simply using the method from the in-domain scenario may also encounter the same problem as in few-shot scenario caused by the domain gap. As a result, we need to find a method to bridge the source domain and target domain. To address this, we propose a combination of the two methods, as outlined in Algorithm \ref{zero-shot template}. In this scenario, we generate different templates for the training set and the testing set, respectively.

First, we select $k$ samples from the training set and testing set, respectively. Then, we compute the weights for the samples in the training set using the Weight-Net. Unlike Algorithm \ref{full-shot template}, we choose the samples in each category with the largest weight as templates to maintain consistency with the testing set. These samples form the template for the training set.

\begin{figure}
\centering 
\includegraphics[width=0.45\textwidth]{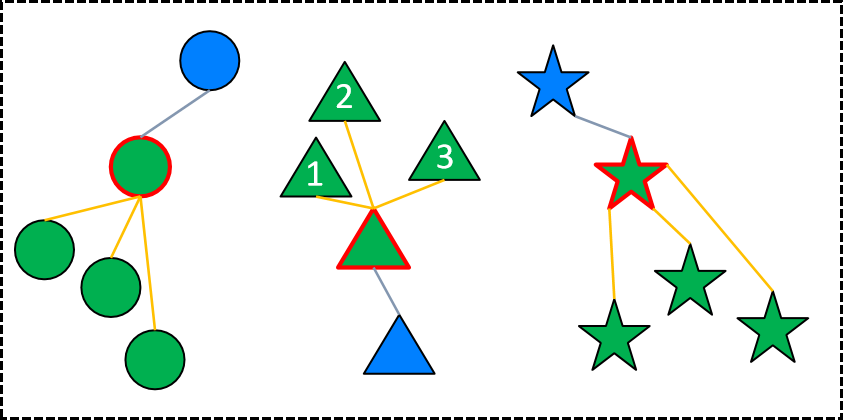}
\caption{Template Generation Method of Target Domain in Zero-shot Scenario: Different shapes represent different categories. The blue items represent the templates from the source domain. The green items represent samples from the target domain. The green items with red lines represent the selected templates of the target domain.}
\label{zero-shot-principle}
\end{figure}

In the inference phase, for the samples in the testing set, we find the most similar ones to the templates of the training set using the similarity scores generated by the CSi-Net. These samples are used as templates for the testing set. The principle of this template generation method is shown in Fig. \ref{zero-shot-principle}. Due to the huge gap between source domain and target domain, directly using templates from the training set would cause a significant performance decrease. For example, in Fig. \ref{zero-shot-principle}, item 1 and 2 would be mistakenly classified as the circle class, and item 3 would be mistakenly classified as the star class.

Inspired by many works that generate pseudo labels for the testing set \cite{EEG}, we take the samples in the target domain that have the highest similarity with templates from the source domain as the templates of the target domain, because they have the highest classification confidence level. Furthermore, many works have proven that samples from the same class and same domain would be easily clustered together in the feature space \cite{tang2020unsupervised}. As a result, the selected templates of the target domain also maintain a high similarity with samples in the same class in the target domain. This method efficiently builds a bridge between templates from the source domain and samples from the target domain.

Although this approach does not guarantee that the samples in the template are correctly categorized, it has a limited impact on the method since different categories also exhibit a certain degree of similarity. We have also demonstrated the effectiveness of this method in Section \ref{Zero-shot Experiment}. Additionally, we can also utilize the optional MK-MMD loss function (Eq. \ref{MK-MMD}) to minimize the gap between the source domain and the target domain as much as possible.

\begin{algorithm}
  \caption{Template Generation in Zero-shot Scenario}
    \textbf{Require}: \\
        \hspace*{1em} $\text{CSi-Net}(\cdot, \cdot)$  \\
	\hspace*{1em} $\text{Weight-Net}(\cdot)$  \\
	\hspace*{1em} Training Set: $C^s$ \\
	\hspace*{1em} Testing Set: $C^t$ \\
        \hspace*{1em} Sample Num: $k$ \\
    \textbf{Return}: \\
        \hspace*{1em} Template for Source Domain: $T^s \in R^{n,2,t,D}$  \\
	\hspace*{1em} Template for Target Domain: $T^t \in R^{n,2,t,D}$  
      \begin{algorithmic} [1]
    \State \Comment{\textbf{Generate Template for Source Domain}}
    \State Sample k CSIs from $C^s$: ${C'}^s \sim \text{Sample}(C^s,k)$
    \State Calculate similarity matrix: $S^s := \text{CSi-Net}({C'}^s, {C'}^s)$
    \State Calculate weights: $W := \text{Weight-Net}(S^s)$
    \State Initialize $T^s\in R^{n,2,t,D}$ and $T_w^s \in R^{n}$ as zero matrices 
    \For{$i = 1$ to $k$}
        \If{$W[i]>T_w^s[\text{label}({C'}^s[i])]$}
            \State $T_w^s[\text{label}({C'}^s[i])]:=W[i]$
            \State $T^s[\text{label}({C'}^s[i])]:={C'}^s[i]$
        \EndIf
    \EndFor
    \State \Comment{\textbf{Generate Template for Target Domain}}
    \State Sample k CSIs from $C^t$: ${C'}^t \sim \text{Sample}(C^t,k)$
    \State Calculate similarity matrix: $S^t := \text{CSi-Net}({C'}^t, T^s)$
    \State Initialize $T^t\in R^{n,2,t,D}$ and $T_w^t \in R^{n}$ as zero matrices 
    \For{$i = 1$ to $k$}
        \State Calculate the category of ${C'}^t[i]$: $y := argmax(S^t[i])$
        \If{$S^t[i,y]>T_w^t[y]$}
            \State $T_w^t[y]:=S^t[i][y]$
            \State $T^t[y]:={C'}^t[i]$
        \EndIf
    \EndFor
    \\ \Return  $T^s,T^t$
  \end{algorithmic}
\label{zero-shot template}
\end{algorithm}

\section{Experiment} \label{Experiment}

\begin{figure}
\centering 
\includegraphics[width=0.45\textwidth]{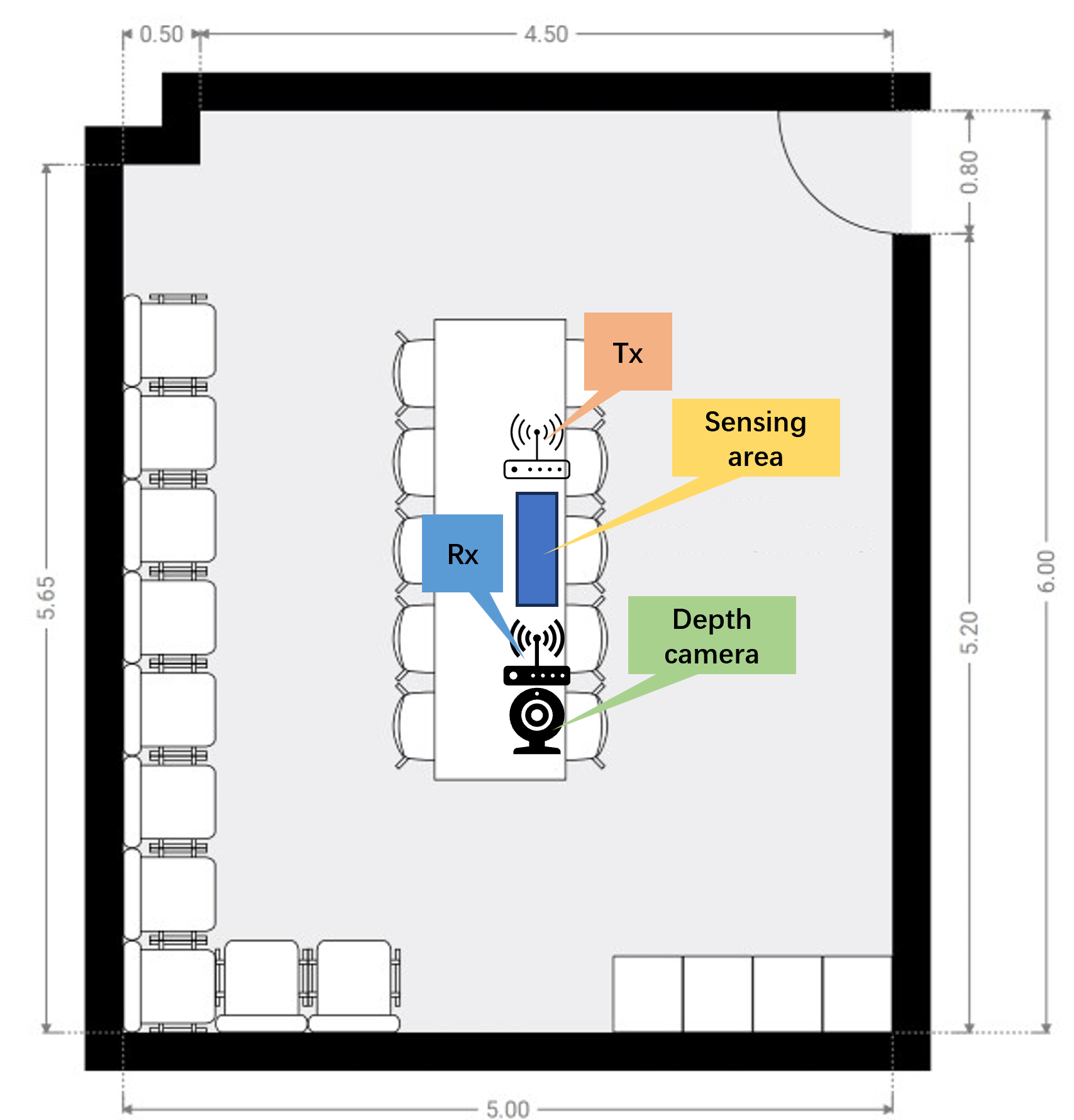}
\caption{Data Collection Environment of WiGesture Dataset.}
\label{env}
\end{figure}

\begin{figure*}
\centering 
\subfloat[Left-right]{\includegraphics[width=0.33\textwidth]{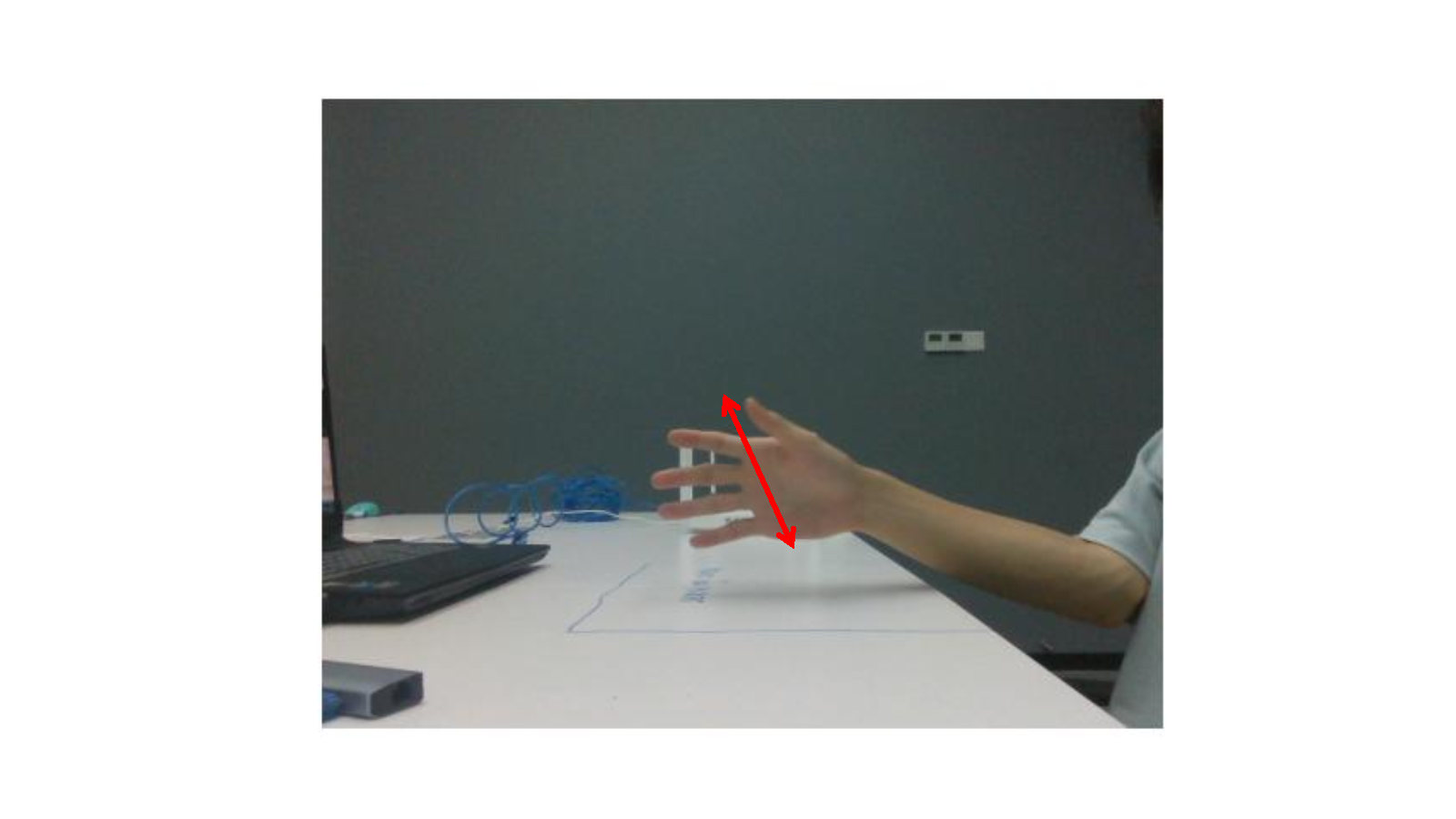}}
\subfloat[Forward-backward]{\includegraphics[width=0.33\textwidth]{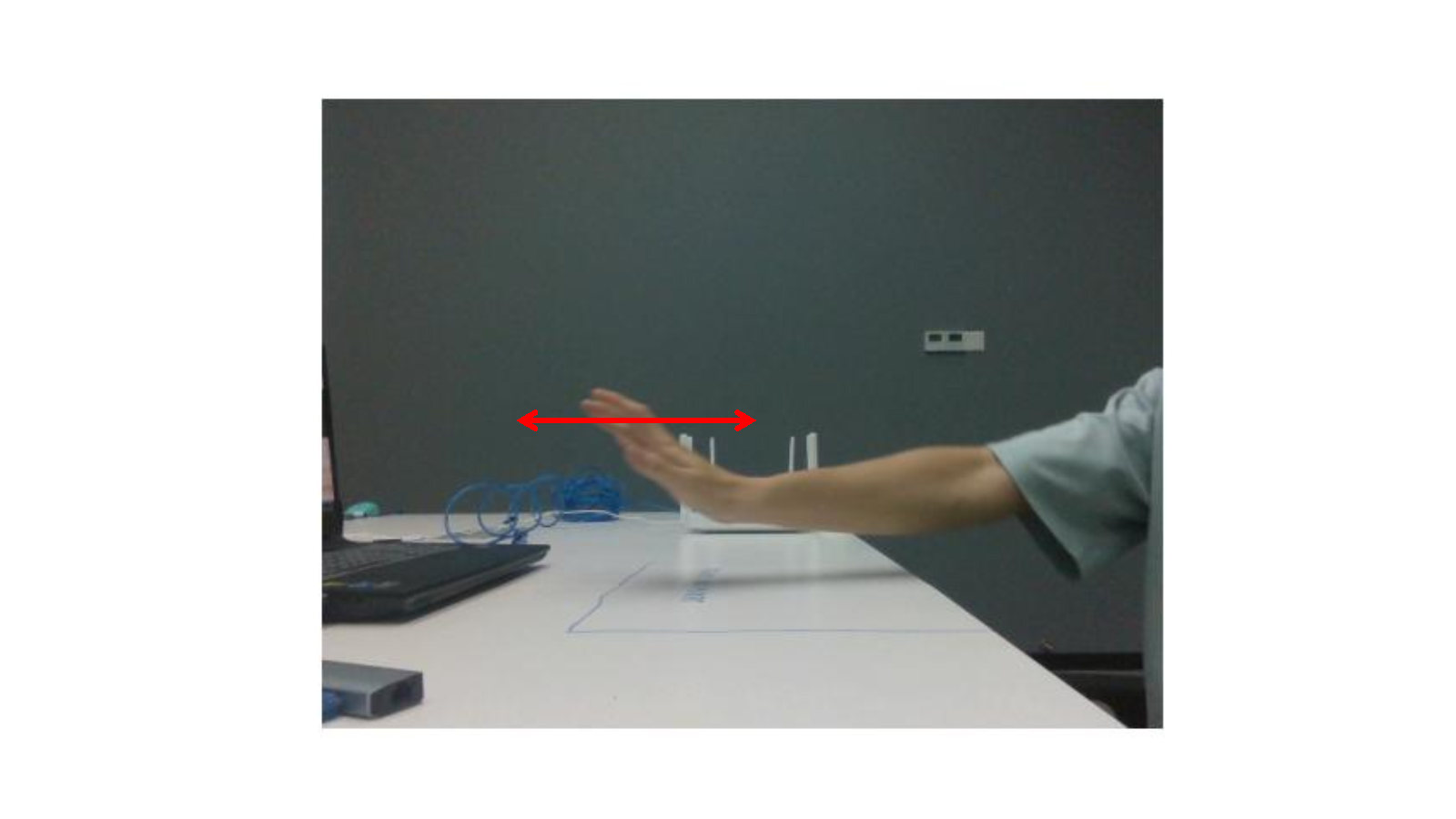}} 
\subfloat[Up-down]{\includegraphics[width=0.33\textwidth]{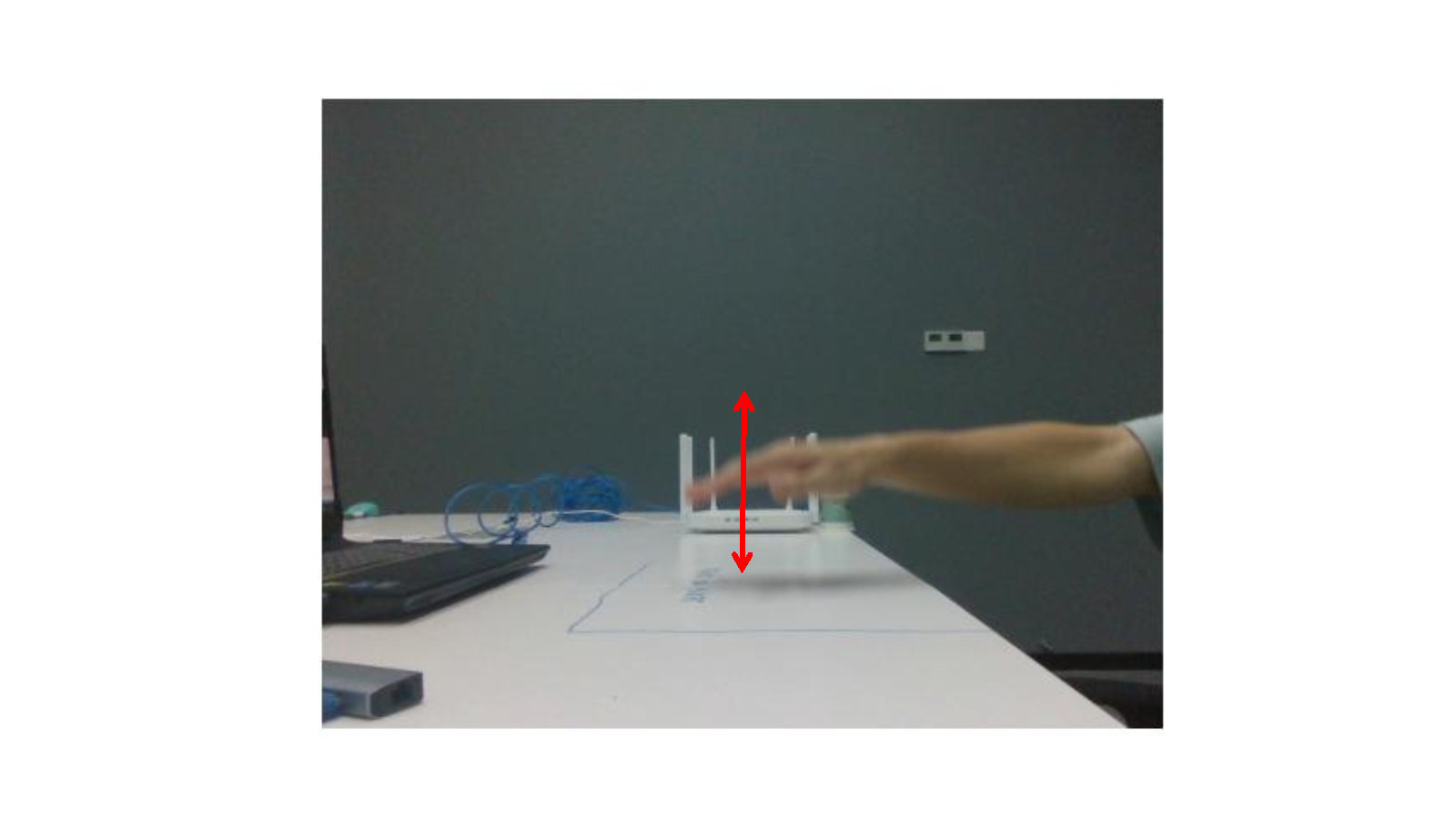}} \\
\subfloat[Circling]{\includegraphics[width=0.33\textwidth]{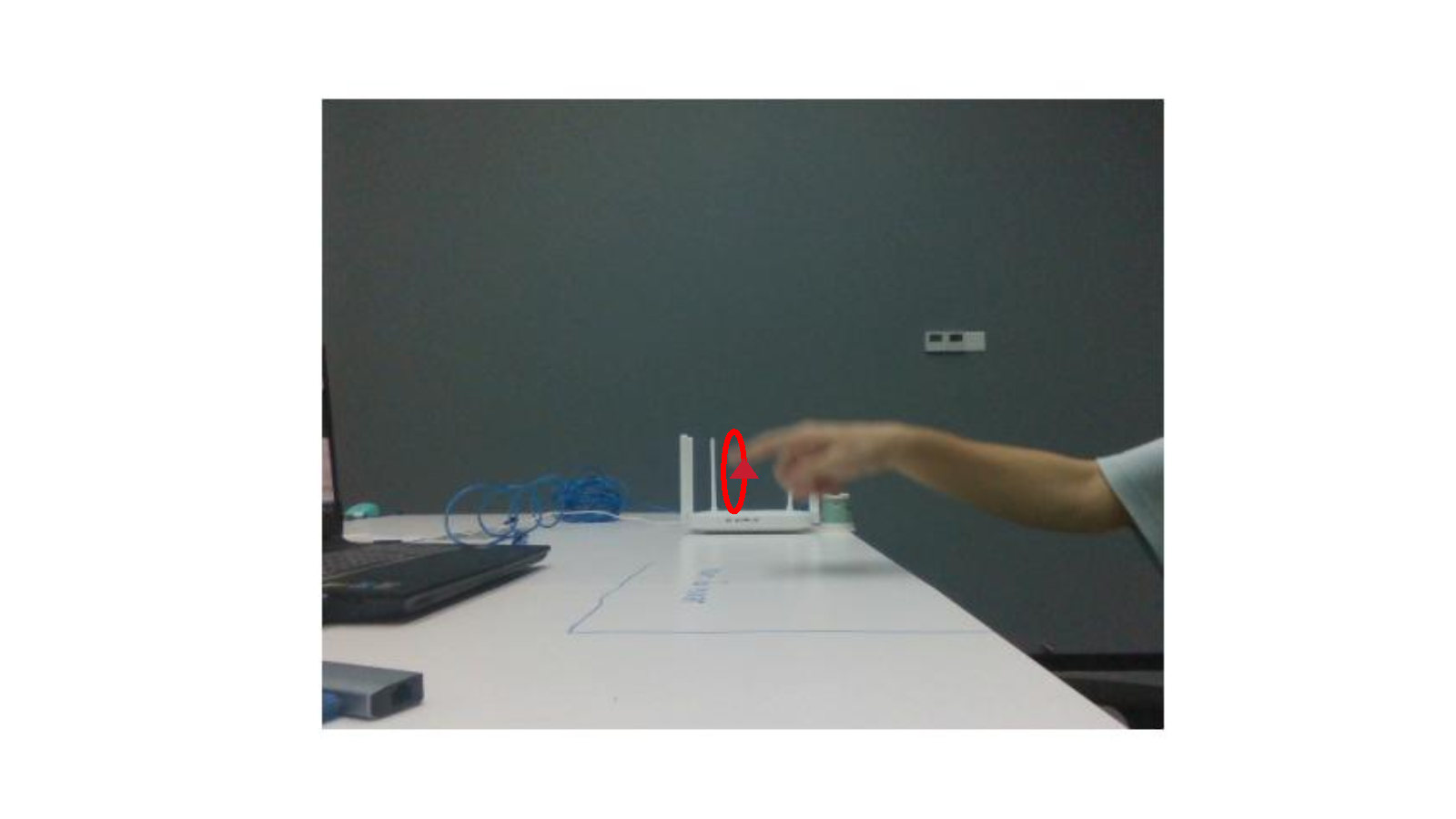}}
\subfloat[Clapping]{\includegraphics[width=0.33\textwidth]{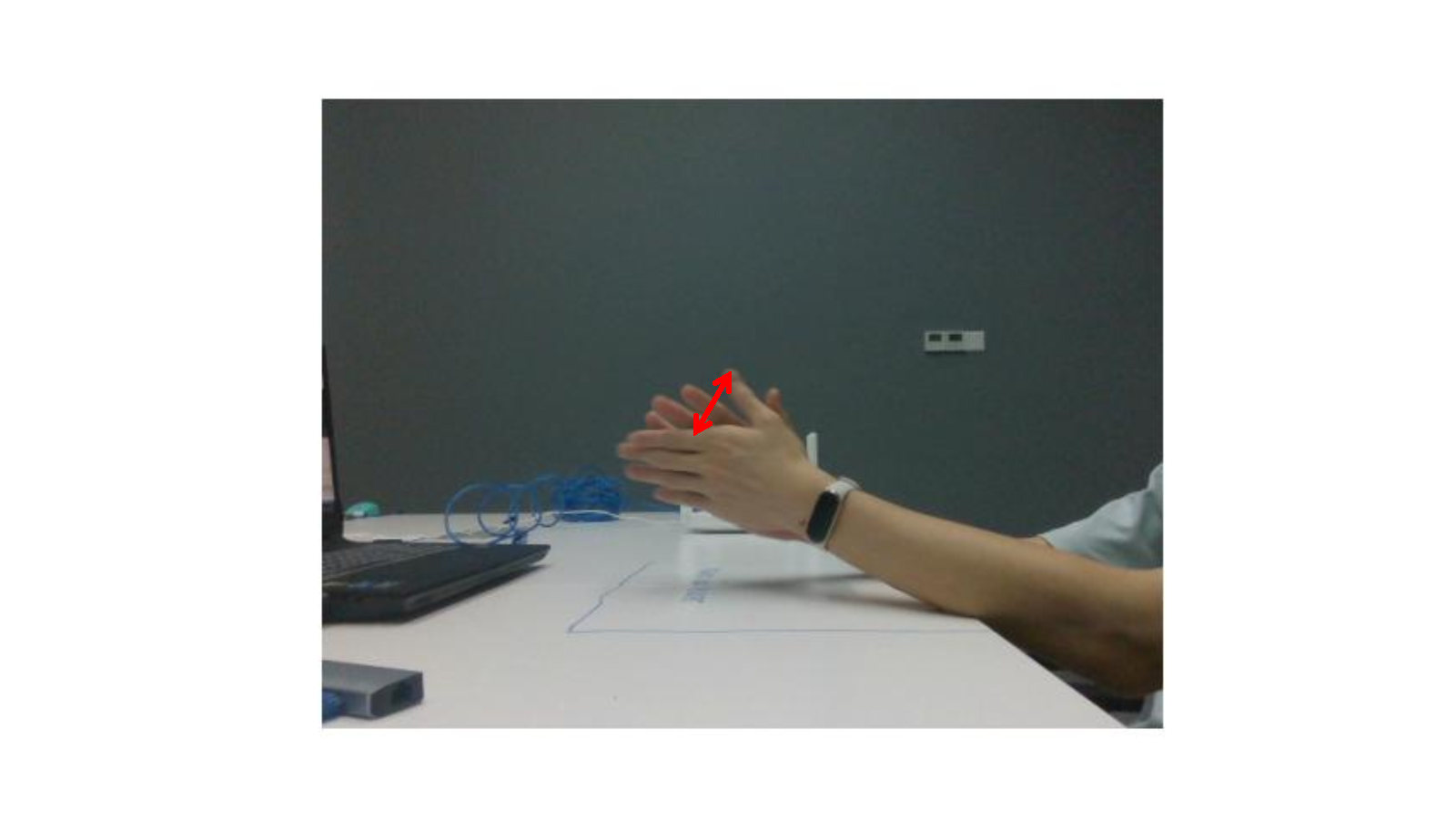}}
\subfloat[Waving]{\includegraphics[width=0.33\textwidth]{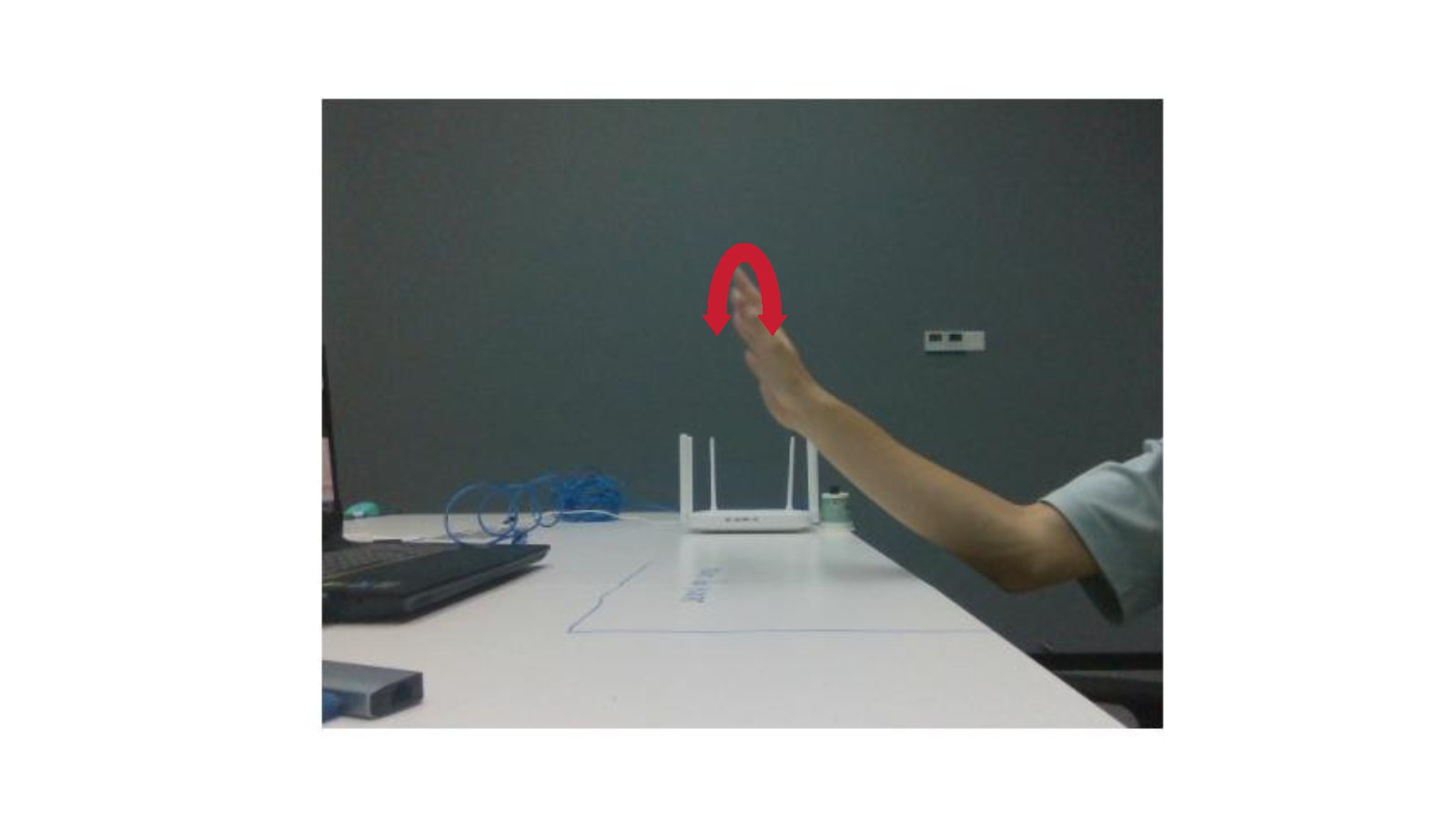}}
\caption{Gesture Sketch Map of WiGesture Dataset \cite{CSI-BERT}}
\label{dataset}
\end{figure*}


\subsection{Experiment Setup}

\subsubsection{Dataset} \label{Dataset}
For the experiments, we utilized the dynamic part of the WiGesture Dataset \cite{CSI-BERT}\footnote{\href{https://paperswithcode.com/dataset/wigesture}{https://paperswithcode.com/dataset/wigesture}}. The dataset was collected using an ESP32-S3 as the RX (receiver) and a home Wi-Fi router as the transmitter. The ESP device is equipped with an antenna and operates at a frequency of 2.4GHz with a sampling rate of 100 samples/s. The data collection environment is depicted in Fig. \ref{env}, with the transmitter and receiver positioned 1.5 meters apart. The dataset consists of a total of 8 individuals who performed various gestures, including left-right, forward-backward, and up-down motions, clockwise circling, clapping, and waving. These gestures are illustrated in Fig. \ref{dataset}.

\begin{figure}
\centering 
\includegraphics[width=0.5\textwidth]{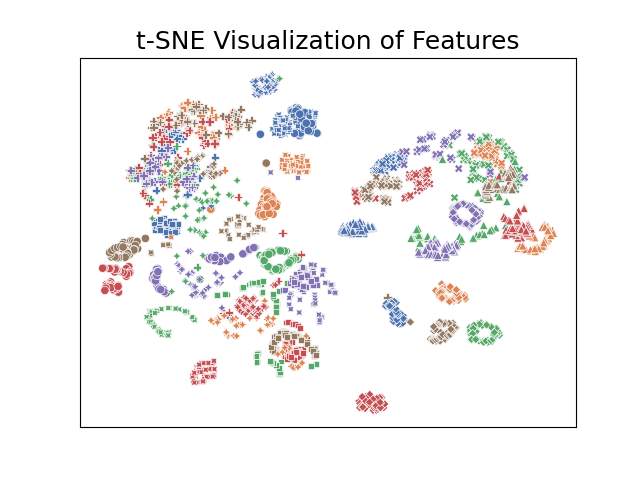}
\caption{t-SNE \cite{tsne} Results for the WiGesture \cite{CSI-BERT} Dataset: Different colors represent different gestures, while different shapes indicate different people.}
\label{tsne}
\end{figure}

To further illustrate the cross-domain challenge present in the dataset, we employ t-Distributed Stochastic Neighbor Embedding (t-SNE) \cite{tsne} to visualize the relationships between samples across different domains. As shown in Fig. \ref{tsne}, we observe that samples corresponding to the same person performing the same action tend to cluster closely together in the feature space. However, when the same action is performed by different people, the samples are often situated far apart, indicating significant variability in how different people execute the same gesture. This observation underscores the complexity of cross-domain recognition tasks, as the model must account for these variations to generalize effectively across different subjects.

\subsubsection{Training Setup}
In our experiments, we implement the CSi-Net and Weight-Net based on ResNet-18 \cite{ResNet}. The total number of parameters in our models is approximately 2.2 million. We observe that our model occupies around 1.5 GB of GPU memory when using a batch size of 64 (in-domain scenario). For optimization, we use the Adam optimizer with an initial learning rate of $5\times 10^{-5}$ and a decaying rate of 0.01. Besides, our experiments are conducted on an Nvidia RTX 3090Ti GPU using the PyTorch framework.


\subsection{In-domain Experiment}
In the in-domain scenario, we use the samples from the first 90\% of the time within each category as the training set, and the remaining last 10\% of the samples as the testing set. We compare CrossFi with previous Wi-Fi Sensing models in the tasks of gesture recognition and people identification. The results are presented in Table \ref{full-shot result}.

\begin{table}
\caption{Experiment in In-domain Scenario: The bold value indicates the best result, which remains consistent across subsequent tables. }
    \centering
    \begin{adjustbox}{width=0.50\textwidth}
        \begin{tabular}{|c||c|c|}
        \hline
        \textbf{Method} & \textbf{Gesture Recognition} & \textbf{People Identification} \\
        \hline 
        \textbf{ResNet-18\cite{ResNet}} & 80.75\% & 86.75\% \\
        \hline
        \textbf{WiGRUNT\cite{WiGRUNT}} & 70.46\% & 97.86\% \\
        \hline
        \textbf{Zhuravchak et al.\cite{LSTM-based}} & 56.93\% & 88.61\% \\
        \hline
        \textbf{Yang et al.\cite{SN-MMD}} & 43.75\% & 87.78\% \\
        \hline
        \textbf{Ding et al.\cite{CNN-based}} & 43.75\% & 61.72\% \\
        \hline
        \textbf{AutoFi (MLP-based)\cite{AutoFi}} & 48.22\% & 89.45\% \\
        \hline
        \textbf{AutoFi (CNN-based)\cite{AutoFi}} & 89.55\%  & 97.74\% \\
        \hline
        \textbf{CSI-BERT\cite{CSI-BERT}} & 74.55\% & 97.92\% \\
        \hline \hline
        \textbf{CrossFi} & \textbf{98.17\%} & \textbf{99.97\%} \\
        \hline
        \end{tabular}
    \end{adjustbox}
\label{full-shot result}
\end{table}

It is evident that our CrossFi achieves superior performance in both tasks, with accuracies exceeding 98\%. Furthermore, we compare CrossFi with its feature extraction component, the ResNet-18 model. We observe that CrossFi outperforms the standalone ResNet-18 by 10\% in terms of accuracy for both tasks.

\subsection{Few-shot Experiment} \label{Few-shot Experiment}
In this section, we evaluate the performance of the models in addressing the challenges posed by cross-domain and new class scenarios, where there are samples belonging to new categories that were not present in the training set. In the cross-domain scenario, we designate gesture with ID 0 and person with ID 0 as the target domain for the gesture recognition and people identification tasks, respectively, while using the remaining data as the source domain. In the target domain, we randomly select $k$ samples from each category as the support set in the k-shot experiment.

\begin{figure*}
\centering 
\subfloat[Gesture Recognition \\ (Cross Domain)]{\includegraphics[width=0.5\textwidth]{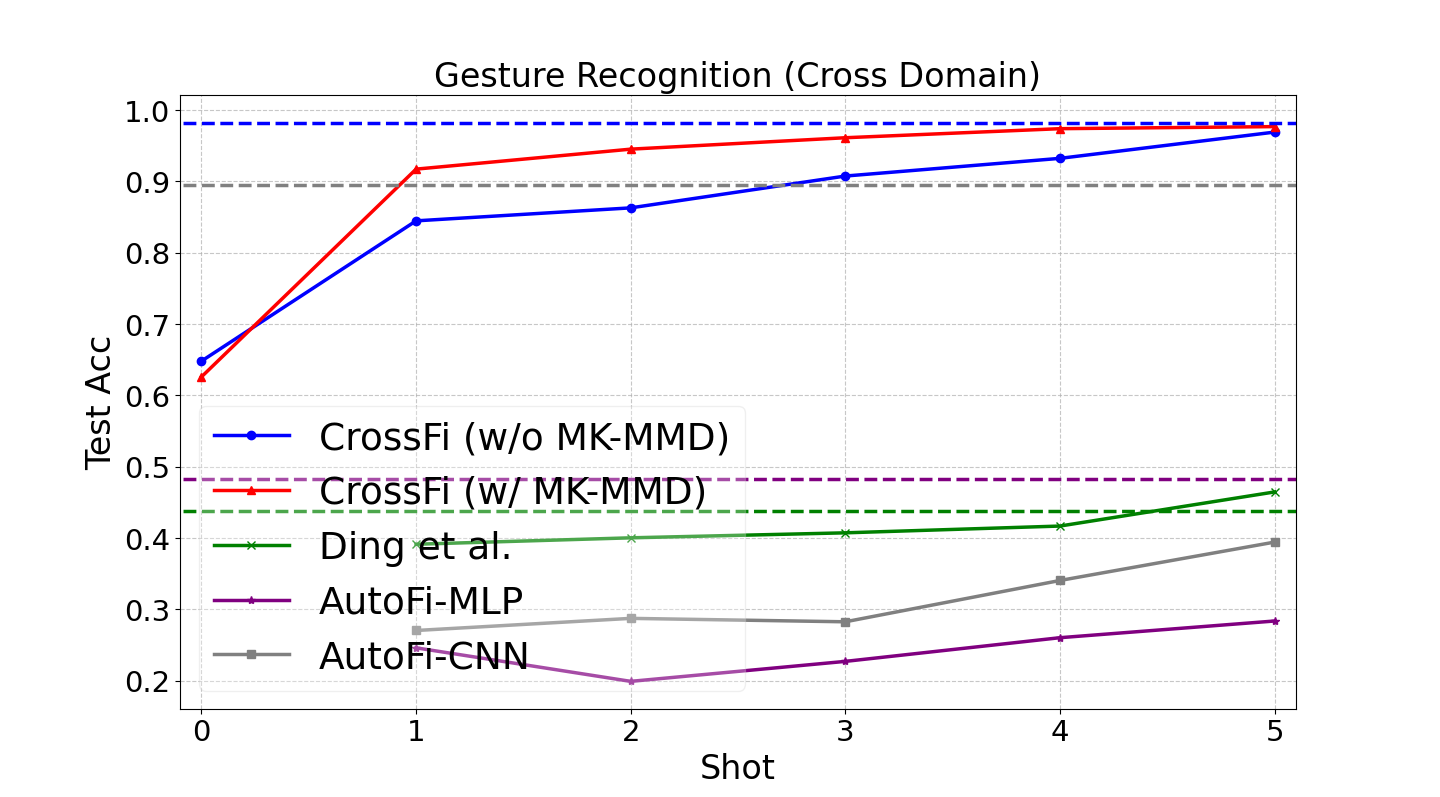}}
\subfloat[People Identification \\ (Cross Domain)]{\includegraphics[width=0.5\textwidth]{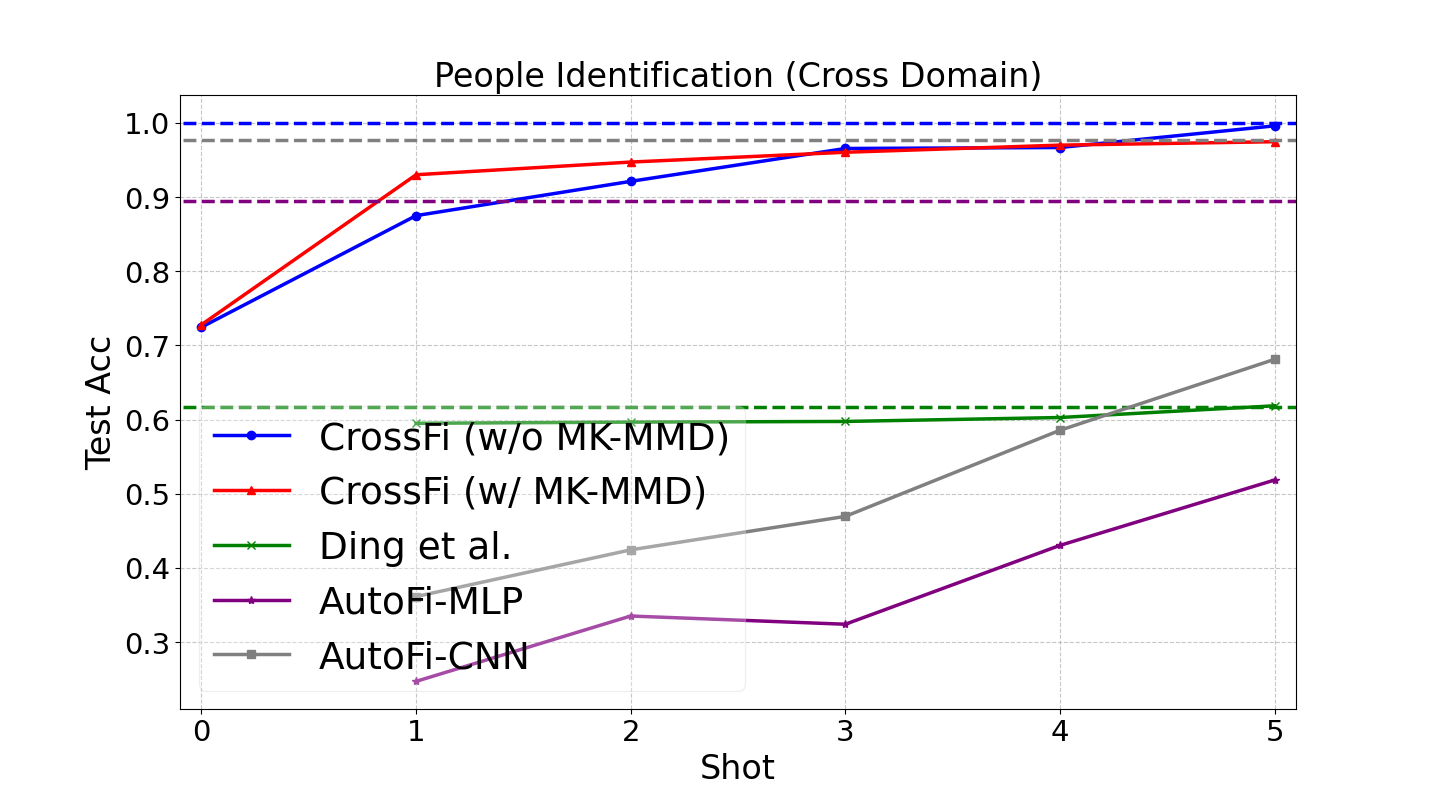}} \\
\subfloat[Gesture Recognition \\ (New Class)]{\label{new-class-gesture}\includegraphics[width=0.5\textwidth]{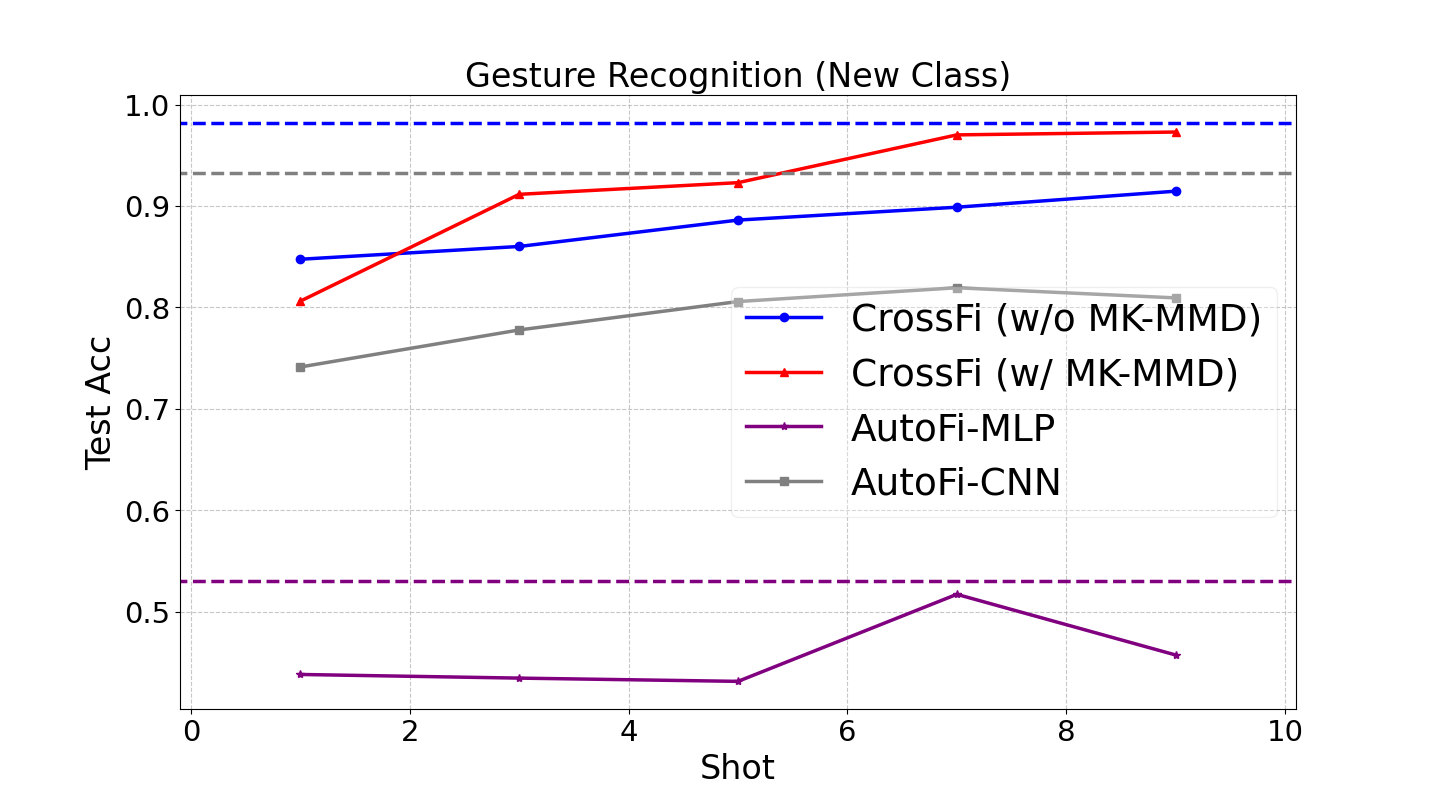}}
\subfloat[People Identification \\ (New Class)]{\label{new-class-people}\includegraphics[width=0.5\textwidth]{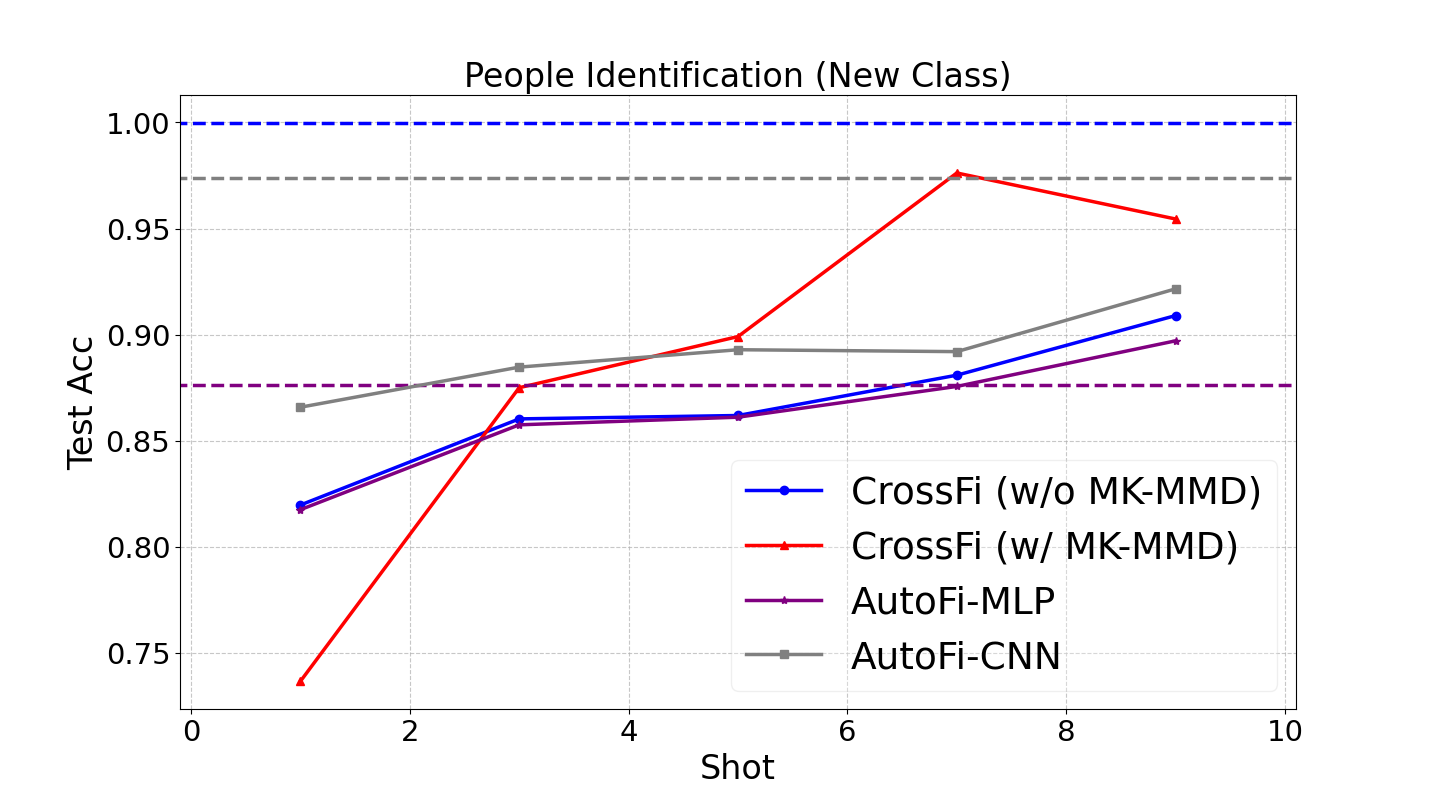}}
\caption{Few-shot Experiment: The figures illustrate the impact of the number of training shots on testing accuracy. The two upper figures represent the cross-domain scenario, while the two lower figures depict the scenario where the testing set contains new classes not present in the training set. In each figure, the dashed lines represent the corresponding model performance in the in-domain scenario.}
\label{few-shot result}
\end{figure*}

Furthermore, in the new class scenario, we utilize all samples from action ID 0 and the last 10\% of samples from other actions as the testing set for the gesture recognition task, with the remaining data serving as the training set. Similarly, for the people identification task, we employ all samples from people ID 0 and the last 10\% of samples from other actions as the testing set, while using the remaining data for training. In this case, we also use $k$ samples for each class from the testing set as the support set in the k-shot scenario. The experimental results are depicted in Fig. \ref{few-shot result}. Specifically, we present the accuracy of the one-shot scenario in Table \ref{one-shot result}. Please note that while there are some methods listed in Table \ref{one-shot result} that can only work in the one-shot scenario or the cross-domain scenario, there are some methods included in Table \ref{one-shot result} that are not shown in Fig. \ref{few-shot result}. 

The results show that CrossFi achieves the best performance in most tasks. Particularly in cross-domain tasks, CrossFi only requires around 5 samples per class to attain accuracy levels comparable to the in-domain scenario.

\begin{table*}
\caption{Experiment in One-shot Scenario.} 
    \centering
    \begin{adjustbox}{width=0.83\textwidth}
        \begin{tabular}{|c||c|c||c|c|}
        \hline
        & \multicolumn{2}{c||}{\textbf{Cross Domain}} & \multicolumn{2}{c|}{\textbf{New Class}} \\
        \hline
        \textbf{Method} & \textbf{Gesture Recognition} & \textbf{People Identification} & \textbf{Gesture Recognition} & \textbf{People Identification}\\
        \hline 
        \textbf{Siamese\cite{siamese}} & 70.40\% & 82.87\% & 66.41\% & 80.92\% \\
        \hline
        \textbf{AutoFi (MLP-based)\cite{AutoFi}} & 24.62\% & 24.71\% & 43.82\% & 81.75\% \\
        \hline
        \textbf{AutoFi (CNN-based)\cite{AutoFi}} & 27.05\% & 36.14\% & 74.13\% & \textbf{86.58\%} \\
        \hline
        \textbf{Yang et al.\cite{SN-MMD}} & 67.21\% & 74.22\% & 58.74\% & 49.00\% \\
        \hline
        \textbf{Ding et al.\cite{CNN-based}} & 39.14\% & 59.50\% & -- & -- \\
        \hline \hline
        \textbf{CrossFi w/ MK-MMD} & \textbf{91.72\%} &  \textbf{93.01\%} & 80.62\% & 73.66\% \\
        \hline
        \textbf{CrossFi w/o MK-MMD} & 84.47\% & 87.50\% & \textbf{84.75\%} & 81.97\%\\
        \hline
        \end{tabular}
    \end{adjustbox}
\label{one-shot result}
\end{table*}

\subsection{Zero-shot Experiment} \label{Zero-shot Experiment}
\begin{table}
\caption{Experiment in Zero-shot Scenario.}
    \centering
    \begin{adjustbox}{width=0.50\textwidth}
        \begin{tabular}{|c||c|c|}
        \hline
        \textbf{Method} & \textbf{Gesture Recognition} & \textbf{People Identification} \\
        \hline 
        \textbf{ResNet-18\cite{ResNet}} & 40.84\% & 70.50\%  \\
        \hline 
        \textbf{ADDA\cite{ADDA}} & 42.71\% & 65.43\%  \\
        \hline 
        \textbf{DANN\cite{DANN}} & 41.41\% & 67.18\% \\
        \hline
        \textbf{MMD\cite{MMD}} & 47.92\% & 67.25\% \\
        \hline
        \textbf{MK-MMD\cite{MK-MMD}} & 40.36\% & 66.47\% \\
        \hline
        \textbf{GFK+KNN\cite{GFK}} & 30.79\% & 51.05\% \\
        \hline \hline
        \textbf{CrossFi w/ MK-MMD} & 62.60\% & \textbf{72.79\%} \\
        \hline
        \textbf{CrossFi w/o MK-MMD} & \textbf{64.81\%} & 72.46\% \\
        \hline
        \end{tabular}
    \end{adjustbox}
\label{zero-shot result}
\end{table}
In the zero-shot experiment, we utilize a similar setup as the few-shot experiment, but without including a support set. The results of the experiment are presented in Table \ref{zero-shot result}. It is important to note that there is limited research on zero-shot learning in the field of Wi-Fi Sensing. Therefore, we have chosen some methods from the domain of machine learning for comparison. Additionally, we use ResNet-18 \cite{ResNet} as a benchmark, which does not include any modules specifically designed for addressing the domain crossing problem. It is evident that our CrossFi significantly outperforms other traditional zero-shot learning methods.

\subsection{Ablation Study}
In this section, we make a series of ablation experiment to illustrate the efficiency of each module of our CrossFi.

\subsubsection{Effect of Target Domain Data Availablility During Training} \label{Effect of Target Domain Data Availablility During Training}
In this section, we prove the efficiency of the MK-MMD loss function and our fine-tuning mechanism in few-shot tasks through two practical scenarios. In the previous discussion, we assumed that during training, we could access a large amount of unlabeled data in the target domain and the full support set. However, these assumptions may not always hold true in practice.

When the unlabeled data in the target domain is not available, we cannot use the MK-MMD to align the distribution between the source domain and the target domain. Even in few-shot scenarios, where there are only a few target domain data points from the support set, it can be challenging to use MK-MMD effectively. This is because MK-MMD relies on calculating the distribution of data from two domains using statistical methods, and with a small amount of data, it can lead to overfitting to the support set. We compare the effect of MK-MMD in Fig. \ref{few-shot result}, which shows that MK-MMD can improve the model's performance. However, our method still works and outperforms most other methods even without MK-MMD. Furthermore, we advise using MK-MMD only when there is a similar label distribution between the source domain and the target domain. During training, the available data in the target domain may be different from the testing set, which means they may have a different label distribution compared to the training and testing sets. In Table \ref{domain MK-MMD experiment}, we demonstrate that when there is a significant difference in labels between the source and target domains, MK-MMD may have a detrimental effect on the model's performance. This phenomenon could be attributed to the distinct distributions of different categories even within a single domain. However, Fig. \ref{new-class-gesture} and Fig. \ref{new-class-people} show that it is beneficial to use MK-MMD to align the new classes with others in new-class tasks. The reason for this can be referred to the explanation of why MK-MMD works in CrossFi, as discussed in Section \ref{few-shot}.

\begin{table}
\caption{Performance of CrossFi (w/ MK-MMD) under Different Circumstances: The experiment is conducted in gesture recognition task, and the label of the training set and testing set are only IDs 2 to 5.}
    \centering
    \begin{adjustbox}{width=0.5\textwidth}
        \begin{tabular}{|c|c||c|c|}
        \hline
        \multicolumn{2}{|c||}{\textbf{Label Distribution in MK-MMD}} & \multicolumn{2}{c|}{\textbf{Testing Accuracy}} \\
        \hline
        \textbf{Source Domain} & \textbf{Target Domain} & \textbf{One Shot} & \textbf{Zero Shot} \\
        \hline
        \multirow{5}{*}{ID 2\~{}5} & ID 2\~{}5 & \textbf{91.72\%} & \textbf{62.60\%} \\
        \cline{2-4}
        & ID 0\~{}5 & 81.39\% & 35.98\% \\
        \cline{2-4}
        & ID 2,3 & 55.04\% & 31.35\% \\
        \cline{2-4}
        & ID 0,1 & 70.74\% & 29.90\% \\
        \cline{2-4}
        & ID 0,2,3 & 69.22\% & 49.16\% \\
        \hline
        \end{tabular}
    \end{adjustbox}
\label{domain MK-MMD experiment}
\end{table}

\begin{figure*}
\centering 
\includegraphics[width=0.8\textwidth]{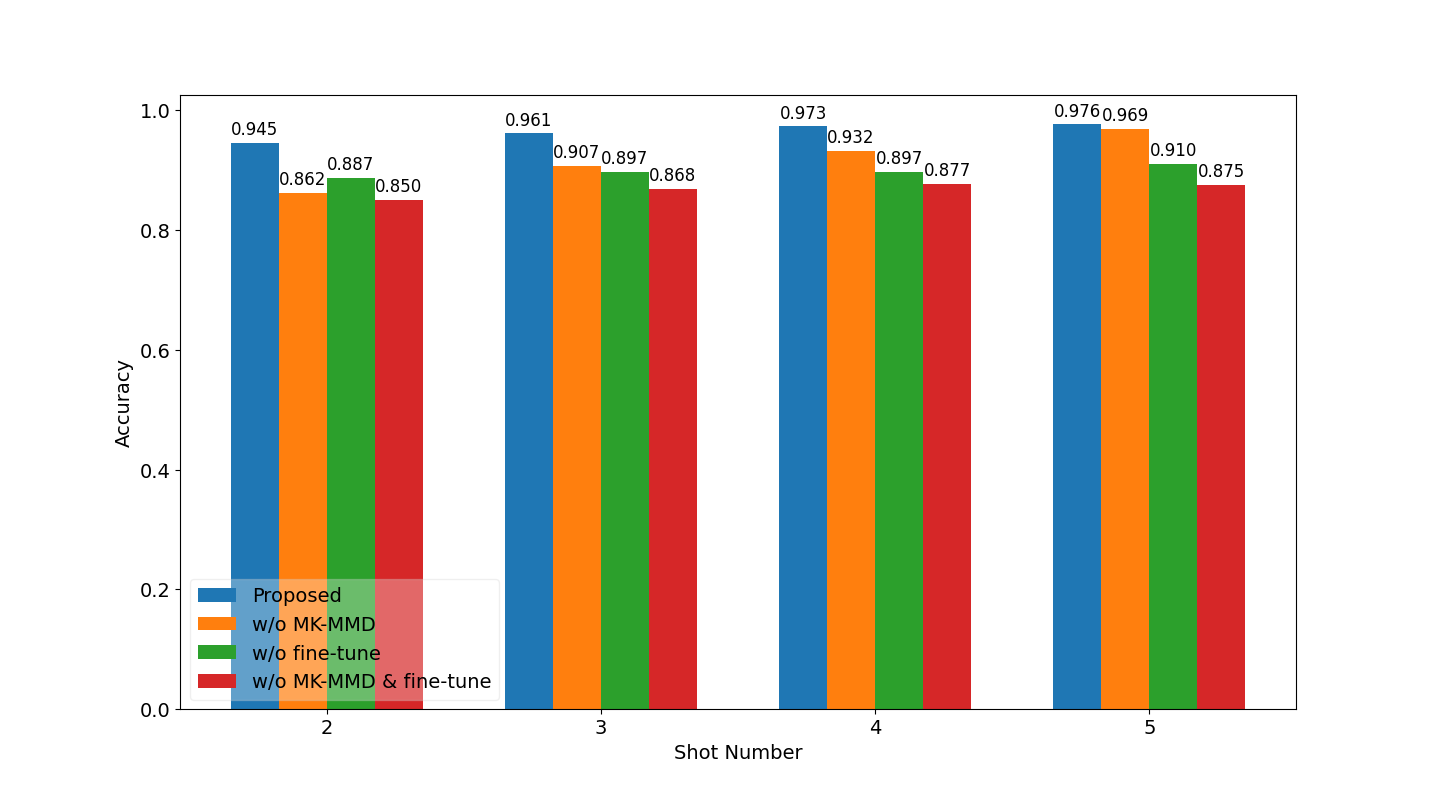}
\caption{Performance of Different Versions of CrossFi in Gesture Recognition Task.}
\label{few-shot ablation result}
\end{figure*}

Moreover, even in few-shot tasks, the support set provides a few labeled data from the target domain. In some practical scenarios, these data are only available during the inference phase, but not during the training phase. In Fig. \ref{few-shot ablation result}, we illustrate the performance of CrossFi without fine-tuning, but being directly applied in the inference phase. The results show that our method can still work even in such a scenario, but it has lower performance compared to the fine-tuning approach. It can also be seen that without fine-tuning, the model performance increases more slowly as the number of shots increases, which suggests the fine-tuning process can efficiently improve the template generation capacity.

\subsubsection{Effect of Multi-Attention Module} \label{Effect of Multi-Attention Module}
In traditional siamese network, the similarity between two samples is typically evaluated using the Gaussian distance between their embeddings. However, we believe that the attention mechanism can provide more informative and suitable similarity evaluation for ``query" and ``key" pairs. To validate this idea, we conduct an ablation study to compare the results of using the attention module versus the Gaussian distance and cosine similarity as the similarity metric.

\begin{table}
\caption{Ablation Study in Similarity Computation Method: The experiment scenario of ``New Class'' is one-shot scenario.}
    \centering
    \begin{adjustbox}{width=0.5\textwidth}
        \begin{tabular}{|c||c|c|c|c|}
        \hline
         \multicolumn{5}{|c|}{\textbf{Gesture Recognition}}  \\
        \hline
         & \textbf{Full Shot} & \textbf{One Shot} & \textbf{Zero Shot} & \textbf{New Class}\\
        \hline 
        \textbf{Gaussian Distance} & 95.58\% & \textbf{84.47\%} & 20.97\% & 74.49\% \\
        \hline
        \textbf{Cosine Similarity} & 91.64\% & 77.17\% & 46.44\% & 74.36\% \\
        \hline
        \textbf{Multi-Attention} & \textbf{98.17\%} & 62.51\% & \textbf{64.81\%} & \textbf{84.75\%} \\
        \hline \hline
         \multicolumn{5}{|c|}{\textbf{People Identification}}  \\
        \hline
         & \textbf{Full Shot} & \textbf{One Shot} & \textbf{Zero Shot} & \textbf{New Class}\\
        \hline 
        \textbf{Gaussian Distance} & 99.74\% & \textbf{87.50\%} & 38.48\% & 80.53\% \\
        \hline
        \textbf{Cosine Similarity} & \textbf{99.97\%} & 83.72\% & 71.16\% & 74.60\% \\
        \hline
        \textbf{Multi-Attention} & \textbf{99.97\%} & 68.04\% & \textbf{72.46\%} & \textbf{81.97\%} \\
        \hline
        \end{tabular}
    \end{adjustbox}
\label{similarity experiment}
\end{table}

According to the results presented in Table \ref{similarity experiment}, the proposed attention measurement method demonstrates superior performance in most scenarios. However, it underperforms in the cross-domain one-shot scenario. Consequently, we have opted to utilize Gaussian distance in the series of experiments involving the WiGesture dataset in the few-shot scenario. It is worth noting that further investigation is warranted to uncover the underlying reasons behind this discrepancy and explore potential avenues for improvement. In the following sections, we will provide a preliminary explanation through experiments.

\subsubsection{Effect of Template Generation Method} \label{Effect of Template Generation Method}
Some previous studies in the field of siamese networks have also employed a similar template generation method to our work. However, they simply compute the average of the training data within each category to create the template \cite{SiFi}, without considering the relationships between different samples and their quality. In this section, we compare the effectiveness of our template generation method with several commonly used methods under the scenario of in-domain and zero-shot.
\begin{table}
\caption{Ablation Study in Template Genration Method.}
    \centering
    \begin{adjustbox}{width=0.5\textwidth}
        \begin{tabular}{|c||c|c||c|c|}
        \hline
         & \multicolumn{2}{c||}{\textbf{Gesture Recognition}} & \multicolumn{2}{c|}{\textbf{People Identification}} \\
        \cline{2-5}
         & \textbf{Full Shot} & \textbf{Zero Shot} & \textbf{Full Shot} & \textbf{Zero Shot} \\
        \hline 
        \textbf{Random} & 94.79\% & 58.83\% & 98.17\% & 60.42\% \\
        \hline
        \textbf{Average} & 91.90\% & 56.39\% & 99.74\% & 68.95\% \\
        \hline
        \textbf{Weight-Net} & \textbf{98.17\%} & \textbf{64.81\%} & \textbf{99.97\%} & \textbf{72.46\%} \\
        \hline
        \end{tabular}
    \end{adjustbox}
\label{template generation experiment}
\end{table}

As depicted in Table \ref{template generation experiment}, the utilization of Weight-Net results in a significant improvement in model performance in both the in-domain and zero-shot scenarios.

\subsection{Expanded Experiment} \label{Expanded Experiment}

\begin{figure*}
\centering 
\includegraphics[width=0.8\textwidth]{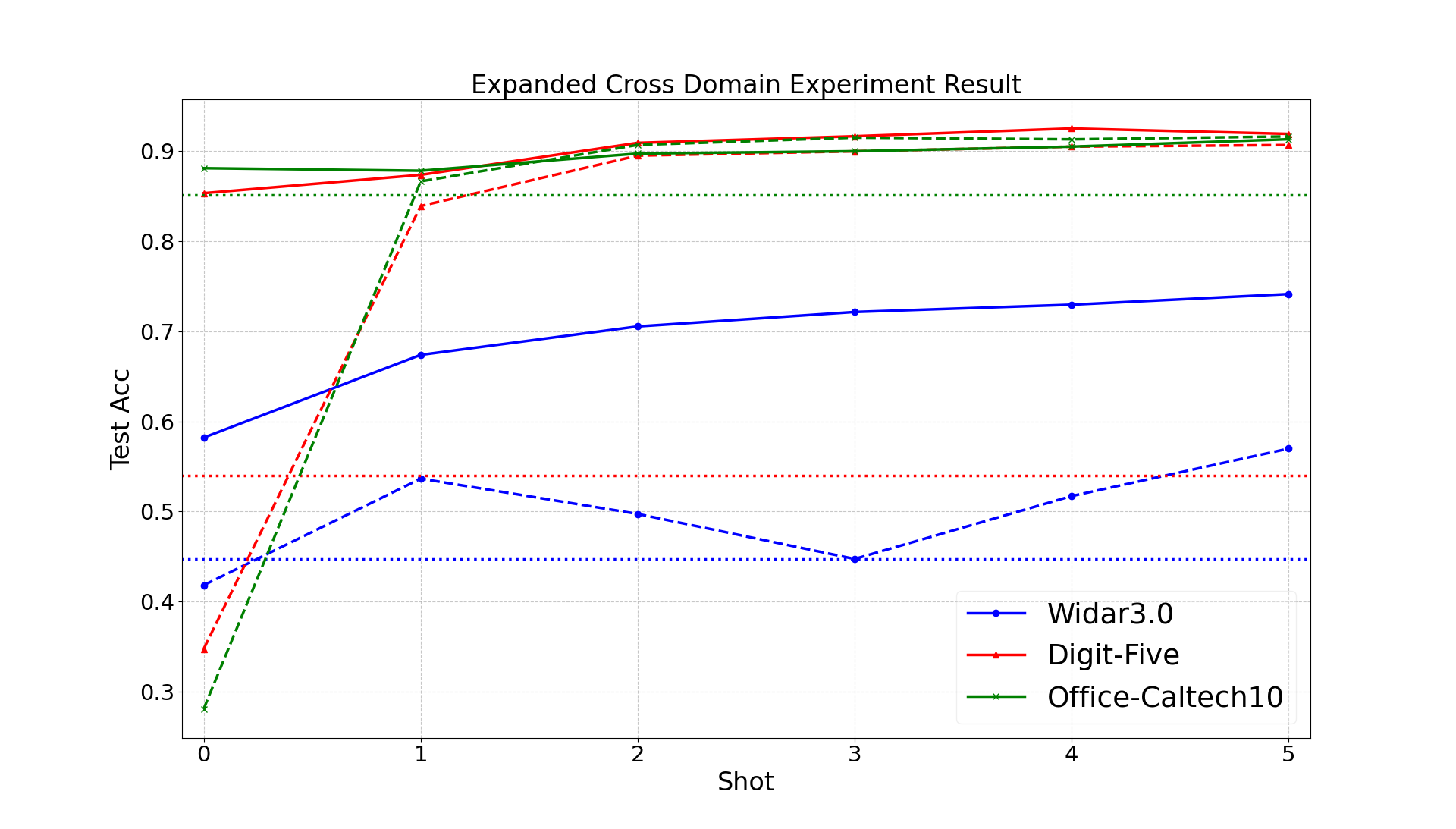}
\caption{Cross-Domain Results in Expanded Datasets: The solid line represents CrossFi with the attention-based similarity calculator, the dashed line represents CrossFi with the Gaussian-based similarity calculator, and the dotted line indicates the benchmark results.}
\label{expand-exp}
\end{figure*}

To further demonstrate the effectiveness of our model, we evaluate its cross-domain performance on additional datasets. Given that CrossFi is not exclusively designed for Wi-Fi sensing, we also conduct experiments on two image classification tasks. The datasets used are as follows:

\begin{itemize}
    \item[$\bullet$] \textbf{Widar3.0 \cite{widar}:} Widar3.0 is a large-scale Wi-Fi activity recognition dataset that encompasses multiple domains, including various individuals and rooms. Our experiments utilize data collected on November 9, 2018, which features six different gestures. We designate User1 as the source domain and User2 as the target domain. The dataset provides processed CSI known as the Body-coordinate Velocity Profile (BVP), with a dimension of $t \times 20 \times 20$, where $t$ represents the time length. Following the approach in \cite{widar}, we pad all samples to $t_{max} \times 20 \times 20$ with zeros for standardization, and subsequently reshape them to $1 \times 20t_{max} \times 20$ for processing by ResNet.

    \item[$\bullet$] \textbf{Digit-Five \cite{digit5}:} Digit-Five is a multi-domain handwritten digit recognition dataset that includes digits from 0 to 9. For our experiments, we select USPS as the source domain and MNIST as the target domain.

    \item[$\bullet$] \textbf{Office-Caltech10 \cite{gong2012geodesic}:} Office-Caltech10 is a multi-domain visual object recognition dataset designed for office environments. It comprises 10 categories, such as books, laptops, and headphones. We choose Amazon as the source domain and Caltech as the target domain.
\end{itemize}

The results of the experiments are illustrated in Fig. \ref{expand-exp}, where the solid line represents CrossFi with the attention-based similarity calculator, the dashed line represents CrossFi with the Gaussian-based similarity calculator, and the dotted line indicates the benchmark result (zero-shot ResNet18). The results indicate that the proposed CrossFi demonstrates excellent performance across various datasets, particularly in Widar3.0 and Digit-Five, where it significantly outperforms the benchmark even in a zero-shot scenario. Additionally, we observe that in these three datasets, the Gaussian-based similarity calculator is not always better than attention-based method, which contrasts with our findings in the WiGesture dataset. Notably, in Widar3.0, the performance of the Gaussian-based method is not even monotonically increasing as the number of shots increases. We believe this is influenced by the extent of domain gap between the source domain and the target domain, which will be discussed in detail in Section \ref{Discussion}.

\section{IoT Deployment} \label{IoT Deployment}
To evaluate the practical applicability of our Wi-Fi sensing approach, we developed a real-time Internet of Things (IoT) system\footnote{A demo is open source at: \href{https://github.com/RS2002/ESP32-Realtime-System}{https://github.com/RS2002/ESP32-Realtime-System}.}. This system is designed to capture and process Wi-Fi signals continuously, enabling real-time monitoring and analysis.

\subsection{Hardware Configuration}

\begin{figure}
\centering
\includegraphics[width=0.3\textwidth]{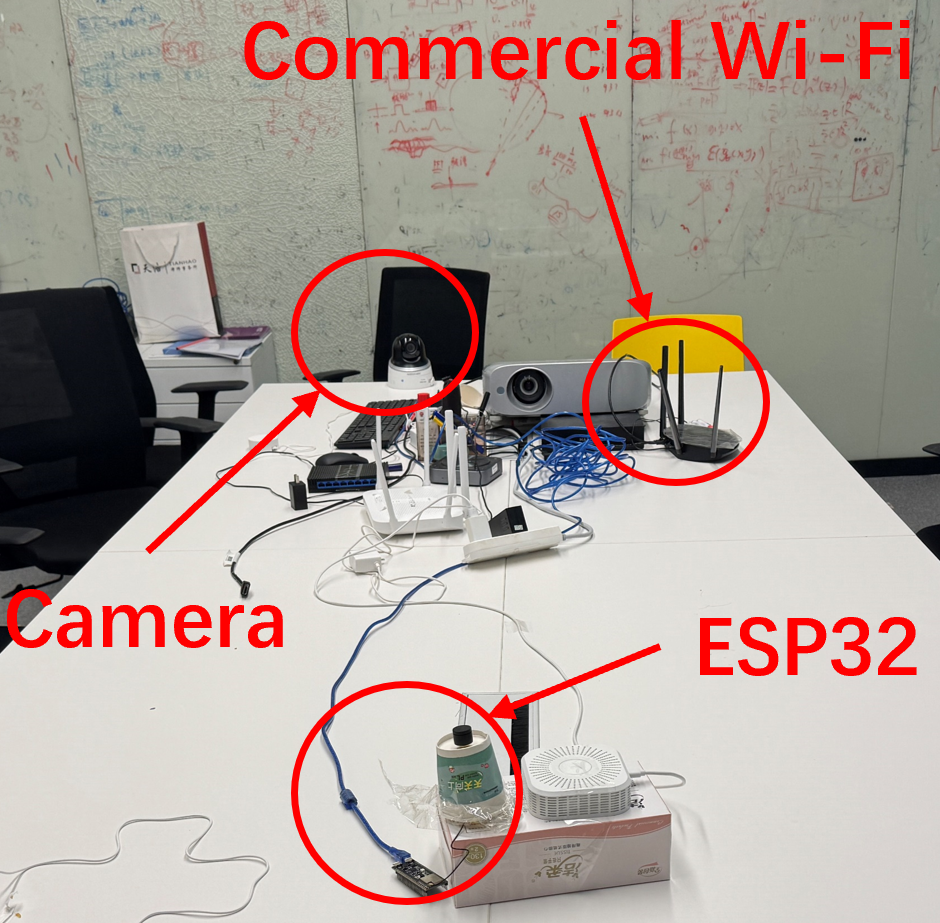}
\caption{Hardware: Our system consists primarily of a commercial Wi-Fi TX and an ESP32 RX. Additionally, a camera is employed to generate labels for the collected data during the training phase, while a remote PC receives the CSI forwarded by the ESP32 and is responsible for downstream calculations.}
\label{fig:hardware}
\end{figure}

Our system, as shown in Fig.~\ref{fig:hardware}, is built around a Personal Computer (PC) that serves as the core unit for both training and deploying the inference models. Equipped with an NVIDIA RTX 3090 GPU, this robust hardware infrastructure supports the real-time processing requirements of our Wi-Fi sensing system, facilitating seamless data analysis and model inference. 

\subsection{System Pipeline}

\begin{figure*}
\centering
\includegraphics[width=\linewidth]{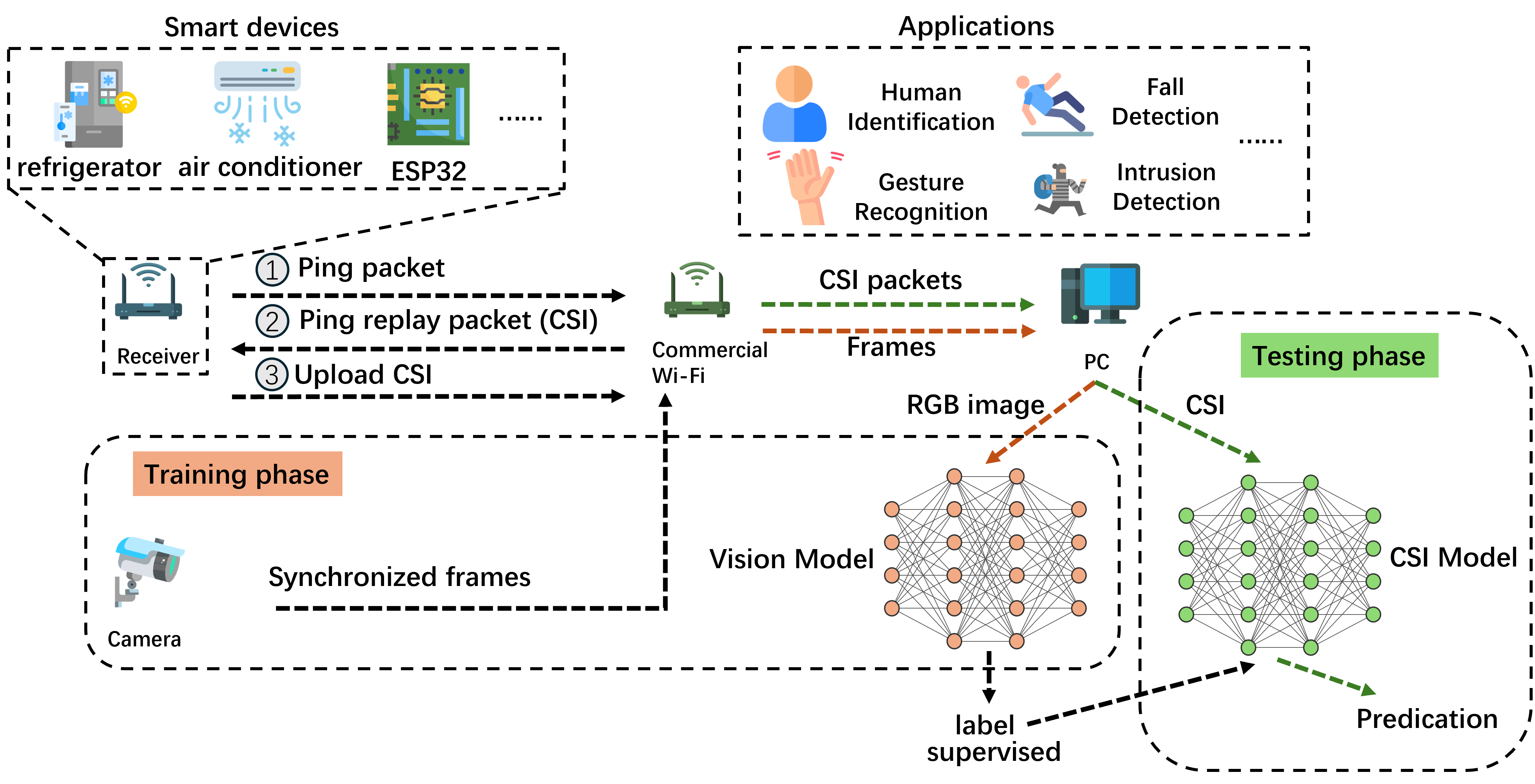}
\caption{System Pipeline: The pipeline consists of two main phases: offline training and online inference. During the training phase, the ESP32 collects CSI while a camera generates labels for the collected samples. The aggregated data is then sent to a remote PC for training. In the inference phase, the ESP32 forwards the received CSI to the remote PC in real-time, allowing the PC to perform calculations based on the trained neural network.}
\label{fig:system_pipeline}
\end{figure*}

The overall pipeline of the system is illustrated in Fig.~\ref{fig:system_pipeline}, encompassing the entire process from data acquisition to inference. The pipeline begins with data collection, where the commercial Wi-Fi router sends ping reply packets to the ESP32-S3, simultaneously capturing CSI data and image frames via the network camera, following \cite{falldewideo}. These data streams are transmitted to the PC for further processing.

During the training phase, the visual model generates corresponding labels from the captured images. These labels supervise the training of the CSI model (i.e. the proposed CrossFi), enabling it to predict environmental changes or object activities based on CSI data. Once the CSI model is adequately trained, the camera can be deactivated, and the system can transition to the inference phase. In this phase, the CSI model independently processes incoming CSI data to produce real-time sensing results without relying on visual inputs.

\subsection{Applications}
Extensive testing has demonstrated that our system operates at a sampling rate of approximately 100 packets per second, which is sufficient to meet the requirements of various sensing tasks. One promising application involves integrating our Wi-Fi sensing system with existing smart home devices, such as refrigerators and air conditioners. By utilizing these devices as CSI receivers, the system can detect gestures or activities within a room and translate them into specific commands. This device-free remote operation offers a novel method for controlling smart home environments without the need for additional sensors, thereby enhancing both flexibility and user experience.

\section{Discussion} \label{Discussion}

\subsection{Complexity and Performance}
In the previous section, we introduced a real-time Wi-Fi sensing system based on CrossFi and discussed its practical deployment in IoT environments, where model complexity and scalability are critical considerations, as they directly impact real-time performance. However, in Section \ref{Expanded Experiment}, we observed a slight performance gap between our model and current state-of-the-art (SOTA) methods in image recognition tasks. A significant factor contributing to this gap is that the feature extractor used in CrossFi is ResNet18 \cite{ResNet}, while most SOTA methods leverage larger models. Therefore, it is essential to strike a balance between model complexity and performance. Since the feature extractor component of CrossFi is not limited to ResNet18, we will compare different feature extractors in this section.

As shown in Table \ref{extractor}, we first compare the performance of various ResNet architectures as feature extractors. We observed that replacing ResNet18 with ResNet34 yields a notable improvement in model performance. However, when we use larger ResNet architectures, we find that model performance does not improve further and may even experience a slight decrease on the WiGesture dataset. One possible explanation for this is that the dataset may not be large enough; when the feature extractor is too large, it becomes challenging to learn the optimal representation of the data.

Next, based on ResNet18, we further explore the impact of using quantization and pruning methods. We found that these techniques can efficiently reduce model size with only a minimal decrease in performance. (It is important to note that the model size and GPU usage between the pruned and original methods remain the same due to the implementation in PyTorch, which masks the pruned parameters rather than deleting them.) As a result, we believe these methods can facilitate practical deployment. Additionally, we observed that even with ResNet101, inference can be completed within one second, processing approximately 350 samples. Thus, directly utilizing our proposed CrossFi method is also practical.

\begin{table*}
\caption{Comparison of Complexity and Performance in One-Shot Cross-Domain Scenario: This method illustrates the feature extractor of CrossFi (with MK-MMD and attention-based similarity calculator). The quantization and pruning versions are based on ResNet18 \cite{ResNet}.}
    \centering
        \begin{tabular}{|c||c|c|c||c|c|}
        \hline
        \textbf{Backbone Model} & \textbf{Model Parameter} & \textbf{Model Size} & \textbf{GPU Occupation} & \textbf{WiGesture \cite{CSI-BERT}} & \textbf{Office-Caltech10 \cite{gong2012geodesic}} \\
        \hline
        \textbf{ResNet34 \cite{ResNet}} & 4.26M & 163.39MB & 1.26GB & 89.18\% & 89.05\% \\
        \hline
        \textbf{ResNet50 \cite{ResNet}} & 4.72M & 180.96MB & 2.36GB & 87.36\%  & 89.91\% \\
        \hline
        \textbf{ResNet101 \cite{ResNet}} & 8.53M & 326.46MB & 3.71GB & 87.33\% & 89.07\% \\
        \hline \hline
        \textbf{ResNet18 \cite{ResNet}} & 2.24M & 85.66MB & 0.93GB & 80.42\% & 87.78\% \\
        \hline
        \textbf{Integer Quantization} & 2.24M & 21.62MB & 0.63GB & 80.36\% & 84.03\% \\
        \hline
        \textbf{Pruning (20\%)} & 1.79M & 85.66MB & 0.93GB & 80.22\%  & 84.81\% \\
        \hline
        \end{tabular}
\label{extractor}
\end{table*}

\subsection{Key Components of CrossFi}
The success of proposed CrossFi can be primarily attributed to its network architecture and training methodology. Comparing to a conventional Siamese network \cite{siamese}, the main innovations of CSi-Net and Weight-Net lie in the attention-based similarity calculator and the adaptive template generator. In this section, we will delve deeper into these two components to further elucidate their functionalities.

\subsubsection{Adaptive Template Generator}
In CrossFi, we propose an adaptive template generation method that utilizes Weight-Net to evaluate the quality of different samples. The quality score obtained from Weight-Net serves as the mixture ratio for template generation, allowing CSi-Net to be expanded into few-shot scenarios. In Section \ref{Effect of Template Generation Method}, we have already demonstrated its superior performance compared to other template generation methods. In this section, we conduct a small experiment to illustrate the efficiency of Weight-Net. After training the model, we select a batch of samples and input them into Weight-Net. We focus on one specific sample within this batch and add Gaussian noise, denoted as $N(0,\sigma^2)$, to it. Table \ref{template} presents the evaluated sample quality scores along with the standard deviation of the added noise. As $\sigma$ increases, the quality score exhibits a downward trend, indicating that the weight of this sample gradually decreases in the template. In this manner, Weight-Net effectively prioritizes high-quality samples over those affected by high noise.

\begin{table*}
\caption{Relationship Between Sample Quality Score and Added Gaussian Noise Standard Deviation}
\centering
        \begin{tabular}{|c||c|c|c|c|c|c|}
        \hline
        \textbf{Gaussian Noise Variance} & \textbf{0} & \textbf{2} & \textbf{4} & \textbf{6} & \textbf{8} & \textbf{10}\\
        \hline
        \textbf{Sample Quality Score} & 0.4690 & 0.4673 & 0.4422 & 0.4474 & 0.4456 & 0.4278 \\
        \hline 
        \end{tabular}
\label{template}
\end{table*}

\subsubsection{Attention-based Similarity Calculator}
In the experiments conducted, we found that the Gaussian-based and attention-based similarity calculators each have their own merits under different tasks. These two methods operate on different principles: the attention-based method functions as a pure black box, relying on neural networks to capture the similarity of samples, while the Gaussian-based method is highly interpretable, relying on the distance between the embedding outputs of samples. Intuitively, the Gaussian-based method depends on the similarity between the source domain and the target domain: only when these domains are similar are samples from the same category more likely to remain close in the embedding space. To demonstrate this, we conducted a series of one-shot experiments in a single source-domain scenario. We used person ID 0 as the source domain and person IDs 4 to 7 as the target domains, utilizing the WiGesture \cite{CSI-BERT} dataset. As shown in Table \ref{similarity experiment}, we compare the performance of the attention-based method with that of the Gaussian-based method. Additionally, to illustrate the similarity between the source domain and the target domain, we used the performance of ResNet18 trained in the source domain as a benchmark. A higher accuracy for the benchmark in the target domain indicates a higher similarity between the source and target domains. We observed that the performance gap between the Gaussian-based method and the attention-based method reflects the same trend as the benchmark performance (i.e., domain similarity). According to our results, we conclude that the attention-based method outperforms the Gaussian-based method when there is a significant gap between the source domain and the target domain. In the next step, we plan to combine the Gaussian-based and attention-based methods to address the problem under varying circumstances, leveraging the strengths of both approaches.

\begin{table}
\caption{One-shot Performance in a Single Source-Domain Scenario: The ``Performance Gap" represents the accuracy difference between the Gaussian-based similarity calculator method and the attention-based method. The ``Benchmark" refers to the zero-shot ResNet18 \cite{ResNet}.}
    \centering
    \begin{adjustbox}{width=0.49\textwidth}
        \begin{tabular}{|c||c|c|c|c|}
        \hline
         \multicolumn{5}{|c|}{\textbf{Source Domain: ID 0}}  \\
        \hline
         \textbf{Target Domain ID} & \textbf{4} & \textbf{6} & \textbf{7} & \textbf{5}\\
         \hline
        \textbf{Multi-Attention} & 94.79\% &	71.57\% &	93.29\% &	65.40\%  \\
        \hline
        \textbf{Gaussian Distance} &  82.18\% &	66.94\% &	96.87\% &	77.34\% \\
        \hline
        \textbf{Performance Gap} &  -12.61\% &	-4.63\% &	3.58\% &	11.94\% \\
        \hline \hline
        \textbf{Benchmark \cite{ResNet}} & 19.51\% &	25.13\% &	31.82\% &	39.92\%  \\
        \hline 
        \end{tabular}
    \end{adjustbox}
\label{similarity experiment}
\end{table}

\subsection{Limitations and Future Directions}
Although our CrossFi framework demonstrates excellent performance across various scenarios and tasks, we acknowledge that it has certain limitations that warrant further exploration in the future. First, in the zero-shot learning scenario, we currently select template samples from the target domains based on their similarity to source domain templates. However, this approach does not guarantee that the chosen target domain templates will be classified correctly. Nevertheless, the method remains efficient: even if a target domain template is misclassified, it still indicates a high degree of similarity to that category, and experimental results have shown promising performance. In the future, we aim to integrate other zero-shot learning methods with ours. If the target domain templates are accurately classified, the performance of the zero-shot learning method should converge with that of one-shot learning. Second, like most domain adaptation methods, our CrossFi currently supports only single target domain scenarios. To broaden its applicability, we can further develop a domain generalization approach based on our framework. Third, as mentioned in the previous section, the attention-based similarity calculation method has room for improvement. We can explore combining its strengths with Gaussian-based methods to enhance model performance further. Last but not least, this paper has demonstrated the model's efficiency in wireless sensing and image recognition tasks. In the future, we can further explore the application of the model in other fields.

\section{Conclusion} \label{Conclusion}
This paper introduces CrossFi, a siamese network-based framework for cross-domain Wi-Fi sensing tasks utilizing CSI. In comparison to traditional siamese networks, we have designed a suitable framework to cater to different scenarios, including in-domain, few-shot, zero-shot cross-domain scenarios, and new-class scenario. Additionally, we have proposed an innovative attention-based method for similarity computation and an adaptively generated template method for the siamese network, which hold potential value for other machine learning domains as well. Through experiments conducted on Wi-Fi gesture recognition and people identification tasks, we have demonstrated that CrossFi achieves superior performance across different scenarios. As our approach is not specifically tailored for Wi-Fi sensing tasks, in the next step, we will evaluate its efficiency in additional fields. Furthermore, we have observed that most models exhibit instability in cross-category scenarios. To ensure practical applicability, further improvements are necessary to enhance the model's stability.


\bibliographystyle{ieeetr}
\bibliography{ref.bib}


\section*{Biography}
\begin{IEEEbiography}[{\includegraphics[width=1in,height=1.25in,clip,keepaspectratio]{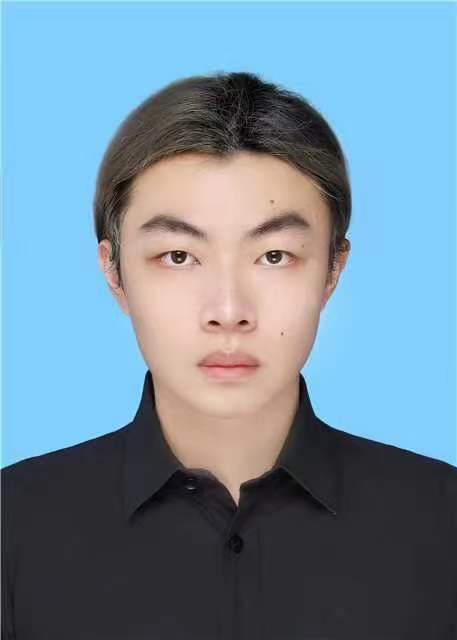}}]{Zijian Zhao}
received the B.Eng. degree in computer science and technology from Sun Yat-sen University in 2024. He is currently pursuing the Ph.D. degree with Department of Civil and Environmental Engineering, The Hong Kong University of Science and Technology. He was a visiting student in Shenzhen Research Institute of Big Data from 2023 to 2024. His current research interests include intelligent transportation, wireless sensing, multi-agent reinforcement learning, and deep learning.
\end{IEEEbiography}

\begin{IEEEbiography}[{\includegraphics[width=1in,height=1.25in,clip,keepaspectratio]{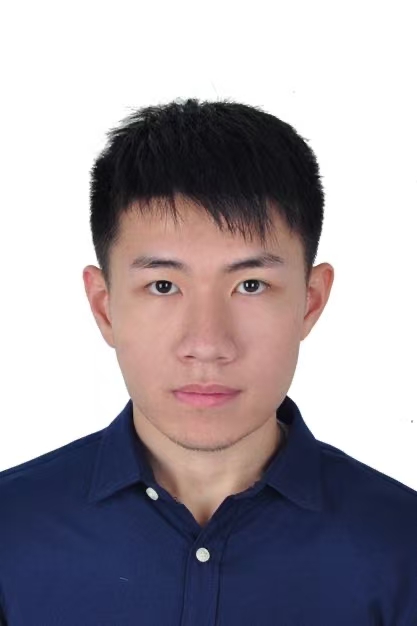}}]{Tingwei Chen}
is currently a visiting student at the Shenzhen Research Institute of Big Data. He received his M.Sc. degree in communication engineering from The Chinese University of Hong Kong, Shenzhen, in 2023 and his B.S. degree in electronic information science and technology from Sun Yat-sen University in 2021. His research interests include multimodal machine learning and wireless sensing.
\end{IEEEbiography}

\begin{IEEEbiography}[{\includegraphics[width=1in,height=1.25in,clip,keepaspectratio]{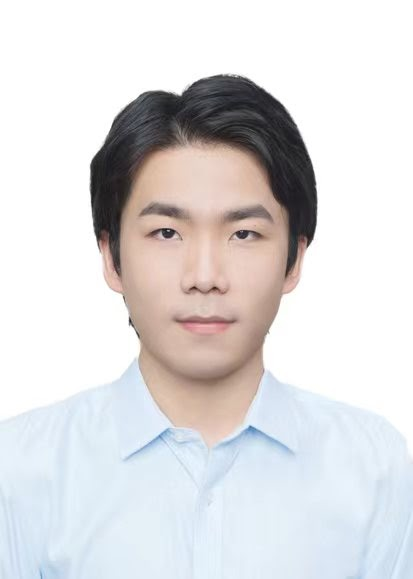}}]{Zhijie Cai}
received the B.Sc. degree in mathematics and applied mathematics from Sun Yat-sen University in 2022. He is currently pursuing the Ph.D. degree with the Shenzhen Research Institute of Big Data, and the School of Science and Engineering, The Chinese University of Hong Kong, Shenzhen. His current research interests include edge intelligence, wireless sensing, and deep learning.
\end{IEEEbiography}

\begin{IEEEbiography}[{\includegraphics[width=1in,height=1.25in,clip,keepaspectratio]{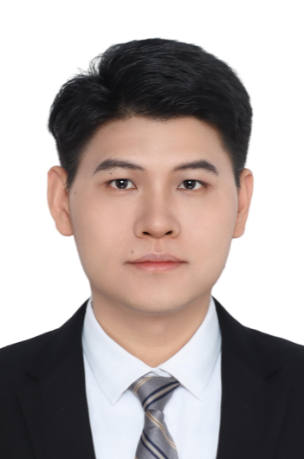}}]{Xiaoyang Li} (Member, IEEE) received the B.Eng. degree from the Southern University of Science and Technology (SUSTech) in 2016 and the Ph.D. degree from The University of Hong Kong in 2020. From 2020 to 2022, he was a Presidential Distinguished Research Fellow at SUSTech. He is currently an Associate Research Fellow with the Shenzhen Research Institute of Big Data, and an Adjunct Assistant Professor with the Chinese University of Hong Kong, Shenzhen. He is a recipient of the Youth Talent Support Program by China Communication Society, Forbes China 30 under 30, Overseas Youth Talent in Guangdong, Overseas High-caliber Personnel in Shenzhen, Outstanding Research Fellow in Shenzhen, the Best Paper Award of IEEE 4th International Symposium on Joint Communications \& Sensing, and the Exemplary Reviewer of IEEE Wireless Communications Letters. He served as the Editor of Journal of Information and Intelligence, the Workshop/Session Chairs of IEEE ICASSP/ICC/WCNC/PIMRC/MIIS. His research interests include integrated sensing-communication-computation, edge learning, and network optimization.
\end{IEEEbiography}

\begin{IEEEbiography}[{\includegraphics[width=1in,height=1.25in,clip,keepaspectratio]{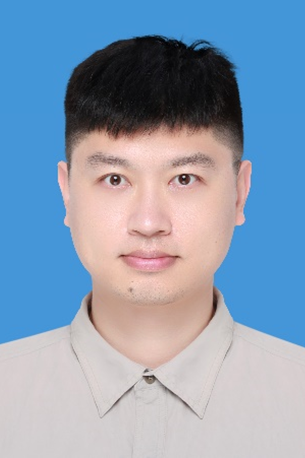}}]{Hang Li}
received the B.E. and M.S. degrees from Beihang University, Beijing, China, in 2008 and 2011, respectively, and the Ph.D. degree from Texas A\&M University, College Station, TX, USA, in 2016. He was a postdoctoral research associate with Texas A\&M and University of California-Davis (Sept. 2016-Mar. 2018). After being a visiting research scholar (Apr. 2018 – June 2019) at Shenzhen Research Institute of Big Data, Shenzhen, China, he has been a research scientist since June 2019. His current research interests include wireless networks, Internet of things, stochastic optimization, and applications of machine learning. He is recognized as Overseas High-Caliber Personnel (Level C) at Shenzhen in 2020. 
\end{IEEEbiography}

\begin{IEEEbiography}[{\includegraphics[width=1in,height=1.25in,clip,keepaspectratio]{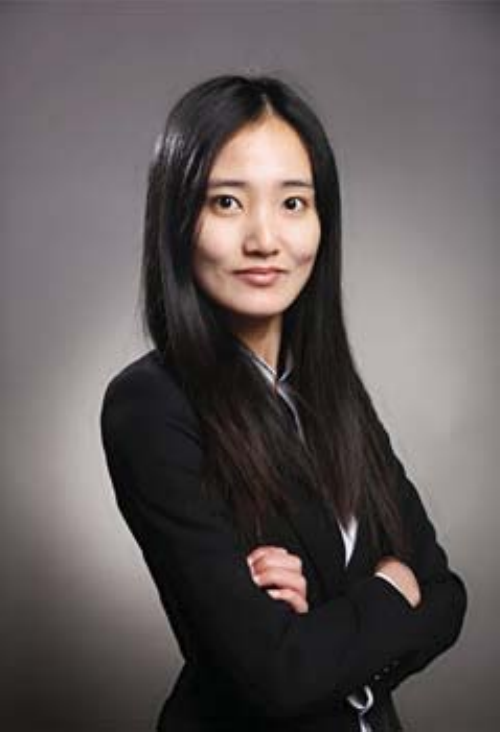}}]{Qimei Chen}
(Member, IEEE) received the Ph.D. degree from the College of Information Science and Electronic Engineering, Zhejiang University, Hangzhou, China, in 2017. She was a Visiting Student with the Department of Electrical and Computer Engineering, University of California at Davis, Davis, CA, USA, from 2015 to 2016. From 2017 to 2022, she was an Associate Researcher with the School of Electric Information, Wuhan University, Wuhan, China, where she has been an Associate Professor, since 2023. Her research interests include intelligent edge communication, unlicensed spectrum, massive MIMO-NOMA, and machine learning in wireless communications. She received the Exemplary Reviewer Certificate of the IEEE Wireless Communications Letters in 2020 and 2023. She has served as a workshop co-chair and TPC Member for IEEE conferences, such as ICC, GLOBECOM, PIMRC, and WCNC. She also served as a Guest Editor for the Special Issues on Heterogeneous Networks of Sensors and NOMA in ISAC (MDPI).
\end{IEEEbiography}

\begin{IEEEbiography}[{\includegraphics[width=1in,height=1.25in,clip,keepaspectratio]{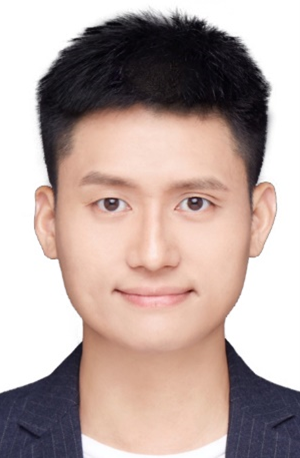}}]{Guangxu Zhu}
(Member, IEEE) received the Ph.D. degree in electrical and electronic engineering from The University of Hong Kong in 2019. Currently he is a senior research scientist and deputy director of network system optimization center at the Shenzhen research institute of big data, and an adjunct associate professor with the Chinese University of Hong Kong, Shenzhen. His recent research interests include edge intelligence, semantic communications, and integrated sensing and communication. He is a recipient of the 2023 IEEE ComSoc Asia-Pacific Best Young Researcher Award and Outstanding Paper Award, the World's Top 2\% Scientists by Stanford University, the "AI 2000 Most Influential Scholar Award Honorable Mention", the Young Scientist Award from UCOM 2023, the Best Paper Award from WCSP 2013 and IEEE  JSnC 2024. He serves as associate editors at top-tier journals in IEEE, including IEEE TMC, TWC and WCL. He is the vice co-chair of the IEEE ComSoc Asia-Pacific Board Young Professionals Committee.
\end{IEEEbiography}

 





\end{document}